\pdfoutput=1

\documentclass{article}

\PassOptionsToPackage{numbers, square}{natbib}

\usepackage[final]{neurips_2024}

\usepackage[utf8]{inputenc} % allow utf-8 input
\usepackage[T1]{fontenc}    % use 8-bit T1 fonts
\usepackage{hyperref}       % hyperlinks
\usepackage{url}            % simple URL typesetting
\usepackage{listings}

\usepackage{booktabs}       % professional-quality tables
\usepackage{amsfonts}       % blackboard math symbols
\usepackage{nicefrac}       % compact symbols for 1/2, etc.
\usepackage{microtype}      % microtypography
\usepackage{xcolor}         % colors
\usepackage{color}
\usepackage{colortbl}
\usepackage{soul}
\usepackage{graphicx}
\usepackage{subfig}
\usepackage{amsmath}
\usepackage{amsthm}
\usepackage{xspace}
\usepackage{soul}
\usepackage[affil-it]{authblk}

\usepackage{algorithm2e}
\usepackage{algorithmic}
\usepackage{thmtools}
\usepackage{enumitem}
\usepackage{pifont}
\usepackage{multirow}
\usepackage{float}
\usepackage{titletoc}
\usepackage{xcolor,soul}
\sethlcolor{lightgray}
\usepackage[titletoc]{appendix}
\usepackage{listings}
\usepackage{adjustbox}

\definecolor{hl}{RGB}{205, 232, 248}

\newcommand*{\affmark}[1][*]{\textsuperscript{\textnormal{#1}}}

\usepackage{tcolorbox}
\newtcbox{\hlwhite}{on line, box align=base, colback=red!20,colframe=white,size=fbox,arc=2pt, before upper=\strut, top=-3pt, bottom=-4.5pt, left=-2pt, right=-2pt, boxrule=0pt}

\definecolor{codegreen}{rgb}{0,0.6,0}
\definecolor{codegray}{rgb}{0.5,0.5,0.5}
\definecolor{codepurple}{rgb}{0.58,0,0.82}
\definecolor{backcolour}{rgb}{0.95,0.95,0.92}
\definecolor{darkpink}{rgb}{0.8, 0.1, 0.5}

\lstdefinestyle{mystyle}{
    backgroundcolor=\color{backcolour},   
    commentstyle=\color{codegreen},
    keywordstyle=\color{magenta},
    numberstyle=\tiny\color{codegray},
    stringstyle=\color{codepurple},
    basicstyle=\tiny,
    breakatwhitespace=false,         
    breaklines=true,                 
    captionpos=b,                    
    keepspaces=true,                 
    numbers=left,                    
    numbersep=5pt,                  
    showspaces=false,                
    showstringspaces=false,
    showtabs=false,                  
    tabsize=2
}
\author{
\textbf{Yushi Hu$^*$}\affmark[1,2]\quad \textbf{Weijia Shi$^*$}\affmark[1]\quad \textbf{Xingyu Fu}\affmark[3]
\quad \textbf{Dan Roth}\affmark[3]\quad \textbf{Mari Ostendorf}\affmark[1] \and\vspace{-6mm}\textbf{Luke Zettlemoyer}\affmark[1]\quad \textbf{Noah A. Smith}\affmark[1,2]\quad \textbf{Ranjay Krishna}\affmark[1,2]
\\
\affmark[1]University of Washington~~~\affmark[2]Allen Institute for AI~~~\affmark[3] University of Pennsylvania 
}

\definecolor{MidnightBlue}{rgb}{0.1, 0.1, 0.44}
\definecolor{gred}{RGB}{234,67,53}
\definecolor{ggreen}{RGB}{52,168,83}
\newcommand{\scoreg}[1]{\color{ggreen}{{#1}}}
\newcommand{\scorer}[1]{\color{gred}{{#1}}}

\definecolor{purp}{HTML}{791f87}

\newcommand{\name}{\textsc{Sketchpad}\xspace}

\title{\includegraphics[scale=0.07]{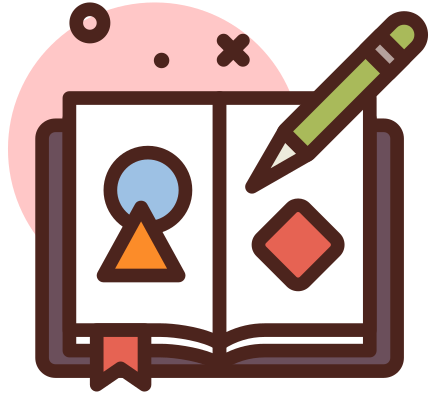}Visual \name: Sketching as a Visual Chain of Thought for Multimodal Language Models}
\begin{document}
\def\thefootnote{*}\footnotetext{Equal Contribution. Correspond to <Yushi Hu: yushihu@uw.edu>, <Weijia Shi: swj0419@uw.edu>}\def\thefootnote{\arabic{footnote}}

\maketitle

\begin{abstract}

Humans draw to facilitate reasoning: we draw auxiliary lines when solving geometry problems; we mark and circle when reasoning on maps; we use sketches to amplify our ideas and relieve our limited-capacity working memory. 
However, such actions are missing in current multimodal language models (LMs). Current chain-of-thought and tool-use paradigms only use text as intermediate reasoning steps. 
In this work, we introduce \name, a framework that gives multimodal LMs a visual sketchpad and tools to draw on the sketchpad. The LM conducts planning and reasoning according to the visual artifacts it has drawn. 
Different from prior work, which uses text-to-image models to enable LMs to draw, \name enables LMs to draw with lines, boxes, marks, etc., which is closer to human sketching and better facilitates reasoning. \name can also use specialist vision models during the sketching process (e.g., draw bounding boxes with object detection models, draw masks with segmentation models), to further enhance visual perception and reasoning. We experiment on a wide range of math tasks (including geometry, functions, graph, chess) and complex visual reasoning tasks. \name substantially improves performance on all tasks over strong base models with no sketching, yielding an average gain of 12.7\% on math tasks, and 8.6\% on vision tasks. GPT-4o with \name sets a new state of the art on all tasks, including $V^*$Bench (80.3\%), BLINK spatial reasoning (83.9\%), and visual correspondence (80.8\%).All codes and data are in \textcolor{darkpink}{\url{https://visualsketchpad.github.io/}}.

% \nascomment{discussion of ``how it works'' is really thin, relative to the lengthy discussion about empirical results.  can you make this more interesting?} Our experiments show that these visual intermediate reasoning steps greatly enhance the model's ability to conduct complex multimodal reasoning. In the math diagram domain, Sketchpad gives 10\% absolute accuracy improvement on geometry problems and 5--20\% improvement on math, graph, and chess problems in IsoBench. On the natural-image domain, GPT-4-turbo gets great improvement on {\it Eyes wide shut} (from 56\% to 71\%), $V^*$ Bench (from 44\% to 71\%), and BLINK (12\% on Relative Depth, 17\% on Jigsaw Puzzles, 10\% on visual correspondence, etc., all absolute) when equipped with our visual sketchpad.  Similar trends are also observed on the stronger model GPT-4o, as well as on open-source models.
\end{abstract}

\section{Introduction}
\label{sec:intro}

\begin{figure}[t]
    \centering
    \includegraphics[width=\linewidth]{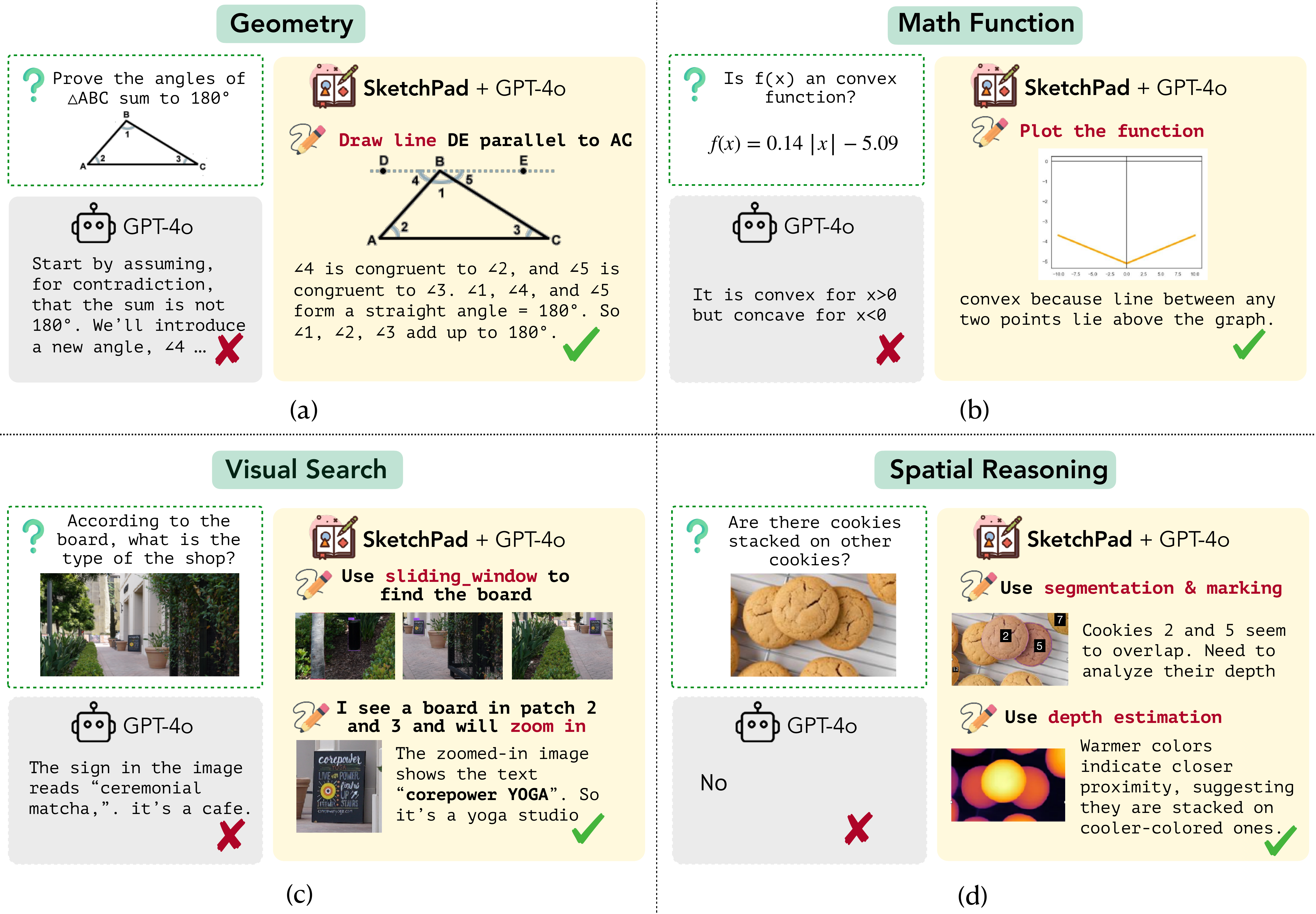}
    \caption{ \textbf{\name equips GPT-4 with the ability to generate intermediate sketches to reason over tasks.}
Given a visual input and query, such as proving the angles of a triangle equal 180°, {\name} enables the model to draw auxiliary lines which help solve the geometry problem. 
The examples are from~\cite{fu2024isobench, wu2023vstar, tong2024eyes}.
For all these examples, without \name, GPT-4o fails to get the correct answer, while \name + GPT-4o achieves the correct solution. 
 } 
\vspace{-0.2in}
    \label{fig:main}
\end{figure}

% SET UP SKETCHING AS A FUNDAMENTAL HUMAN ACTIVITY IF HUMANS ATTEMPTS THESE TASKS.
Sketching is a fundamental human activity, serving as a versatile tool for communication~\cite{goel1995sketches}, ideation~\cite{tversky2003sketches}, and problem-solving~\cite{tversky2009thinking}. 
Unlike written language, sketches have the advantage of conveying visuo-spatial ideas directly, for example by using spatial relations on paper to convey spatial relations or other more abstract relationships in the world. This may explain their ubiquity; maps~\cite{taylor1992spatial} and architectural plans~\cite{goldschmidt1991dialectics} have been found incised in stone, etched on leather, impressed in clay, and drawn on paper in diverse cultures scattered across the world~\cite{taylor1992descriptions}. 
%Sketches serve as a form  of \textit{cognitive scaffolding}: people use sketches to amplify their ideas and relieve their limited-capacity working memory.
% From a neuroscience perspective, sketches also call to action our \textit{place cells} or \textit{grid neurons}~\cite{tversky2019mind}, involving more cognitive support to reason.
Sketches are so fundamental that we use them to teach school children how to solve geometry problems by drawing support lines, to aid engineers conveying prototypes, to support architects creating blueprints, and to allow scientists like us to convey scientific contributions (see Figure~\ref{fig:main}).

% TALK ABOUT THE KINDS OF TASKS WE WANT MULTIMODAL LMS TO BE ABLE TO DO. END BY SAYING THAT THEY SUCK RIGHT NOW.
% In these benchmarks, models are given an image of a geometric shape with annotations in the image itself and asked a question that requires symbolic grounding and spatial understanding. 

As multimodal language models (LMs)~\cite{gpt4, team2023gemini, liu2023llava, liu2024llavanext, alayrac2022flamingo, bai2023qwenvl, wang2023cogvlm, dai2023instructblip, lu2023unified, Team2024ChameleonME, chen2023internvl, chen2023palix} have begun to mature, we now expect them to solve tasks like the ones mentioned above, i.e., ones where people draw intermediate sketches to simplify reasoning.
Popular benchmarks now include questions about geometry (e.g., Geometry3K~\cite{lu2021inter}) and complex math problems (e.g., IsoBench~\cite{fu2024isobench}). 
In these benchmarks, models are given images of diagrams and asked questions requiring symbolic grounding and spatial understanding, where intermediate sketches like auxiliary lines can enhance reasoning.
Even benchmarks in computer vision now have a similar flavor.
Specialist vision models can be viewed as sketching on natural images. For example, object detection is plotting bounding boxes around objects; depth estimation is drawing colormaps according to depth. 
The recently proposed BLINK benchmark~\cite{fu2024blink} would benefit significantly from such intermediate visual sketches. Similarly, the \textit{V*}Bench benchmark~\cite{wu2023vstar} could focus reasoning on image crops to find answers.
% requires models to answer questions that become trivial if the model could produce intermediate sketches annotating the input image with depth estimates or match corresponding keypoints between multiple input images. MMVP~\cite{tong2024eyes} and \textit{V*}Bench~\cite{wu2023vstar} are other benchmarks where sketching to focus reasoning on image crops can simplify finding answers.
% Despite the potential for sketch-based reasoning across math and computer vision, 
Unfortunately, current LMs lack a scaffold for using sketch-based reasoning when solving tasks. 
% \nascomment{reviewer is now wondering if we will evaluate on these benchmarks.  if we don't, they will be disappointed.  maybe add a sentence here that says explicitly that this paper will show how a visual sketchpad helps with the evaluations captured by these specific benchmarks, if it's true} 
% \weijia{not sure if we should add one sentence in this paragraph as the current version seems to flow well?}

% OUR CONTRIBUTION:
In this paper, \textbf{we introduce Visual \name: a framework that provides multimodal LMs with the tools necessary to generate intermediate sketches to reason over tasks.} Inspired by textual chain-of-thought reasoning in LMs \citep{wei2022chain, zhang2023multimodal}, \name prompts the underlying visual LM to produce visual artifacts as part of a chain of mixed textual, programmatic, and visual reasoning. 
For example, to prove that the angles of triangles sum up to 180 degrees in Figure~\ref{fig:main} (a), \name enables agents to modify the diagram by introducing a new auxiliary line. This new line, along with new annotated angles, provides the critical information to solve the geometry task. Similarly, \name improves models' spatial reasoning for computer vision. To determine if there are cookies stacked on top of other cookies in the image (Figure~\ref{fig:main}b), the model first produces an intermediate depth estimate. 
% Since the depth estimation produces intersecting cookies at different depths, the model is able to correctly answer that yes, the cookies are stacked.
By analyzing the depth estimate, which reveals cookies overlapping at different depths, the model is able to correctly answer that the cookies are indeed stacked.

% OUR EXPERIMENTS WITH MATH:
We demonstrate the effectiveness of visual \name across a wide range of mathematics and computer vision tasks. For math, we tackle problems including (1) geometry ~\citep{lu2021inter}, (2) mathematical functions, (3) graph algorithms, and (4) strategy games ~\citep{fu2024isobench}. For geometry questions, \name enables models to generate Matplotlib code with auxiliary lines and variables, given the diagram input and questions (\autoref{fig:main}a). Notably, even when the input is solely language-based, such as mathematical functions, \name enables models to plot the functions and reason about their properties, using only the mathematical function expression as input (\autoref{fig:main}b). These results highlight the ability of \name to aid reasoning, even in tasks with purely language-based inputs.
\textbf{Across all four categories of mathematical tasks, \name consistently improves the baseline GPT-4o performance, yielding an average gain of 11.2\%.} 
% OUR EXPERIMENTS WITH VISION:
For computer vision, we tackle diverse tasks including (1) depth, (2) spatial reasoning, (3) jigsaw, (4) visual correspondence, (5) semantic correspondence, as well as questions from (6) the MMVP  and (7) the \textit{V*}Bench benchmarks~\cite{fu2024blink, tong2024eyes, wu2023vstar}. For this domain, \name enables models to generate segmentation masks, crop images, draw bounding boxes, zoom into image regions, overlay images, etc. Similar to math, \textbf{\name shows consistent improvements across all seven types of computer vision tasks}. For example, GPT-4o, augmented with \name, sees 14.3\% improvement on \textit{V*}Bench, 12.1\%, and 9.7\% improvements on BLINK's depth and semantic correspondence tasks, setting  a new state of the arts across all tasks.
Finally, we analyze the effectiveness of \name{} by comparing the plans generated by our model with human-created plans, showing that they are well-aligned and exhibit similar reasoning patterns. 
%We also show that sketches like diagrams and plots can facilitate open-source multimodal LMs like LLaVA~\cite{liu2024llavanext}.
We hope \name opens up new research opportunities toward more capable and interpretable multimodal intelligence.

\vspace{-0.1in}
\section{Related Work}
\vspace{-0.1in}

\label{sec:related_work}
\name generalizes recent work on multimodal tool-use and visual prompting. We also place our work within the larger sphere exploring LMs as agents.

\noindent\textbf{Visual programming and tool-use.}
With the advancement of LMs~\cite{brown2020language, gpt4,team2023gemini,llama2, groeneveld2024olmo}, researchers have demonstrated the possiblity of decomposing complex vision tasks into simpler substeps that can each be solved using vision tools~\cite{yang2023mmreact, zeng2022socratic,hu2023visual,hu2022promptcap}. Among them, the most relevant to us are Visprog~\cite{gupta2023visual} and ViperGPT~\cite{surismenon2023vipergpt}. They use LMs to generate Python code, which sequencially invokes  specialized vision tools. 
These methods share a common problem that the multimodal modules follow a pre-defined plan outlined by the LM. By contrast, \name allows LMs to change their plan according to the intermediate visual artifacts, yielding better performance and robustness when solving complex multimodal tasks.

\noindent\textbf{Visual prompting.}
Recent work shows that multimodal models can be augmented by visual prompts added to natural images~\cite{shtedritski2023does}. For example, SoM~\cite{yang2023setofmark} shows that adding labeled segmentation masks on images unleashes GPT-4V's visual grounding ability. Prior work also reports similar findings in 3D~\cite{liu20233daxiesprompts} and Robotics~\cite{google2024pivot}. \name is a generalized framework for all these methods, allowing LMs to decide what visual prompting to use as part of the multimodal reasoning process.

\noindent\textbf{LMs as agents.}
Recent work has started to treat LMs as agents that can both reason and act~\cite{yao2022react,ma2024m,wu2023autogen, shi2023replug, yasunaga2023retrieval}. Researchers have applied this idea to software engineering~\cite{jimenez2023swe,zhang2023ecoassistant,hong2023metagpt}, robotics~\cite{google2024pivot}, vision~\cite{liu2023llavaplus, yang2023mmreact}, and GUI navigation~\cite{yan2023gpt, koh2024visualwebarena, xie2024osworld}. \name can also be viewed as an agent that accepts multimodal inputs and outputs. One big difference is that \name can create visual artifacts to facilitate reasoning, while prior LM agents only generate text during reasoning.

\vspace{-0.05in}
\section{Visual \name}
\vspace{-0.05in}

\label{sec:method}

\begin{figure}[t]
    % \left
    \includegraphics[width=1.01\linewidth]{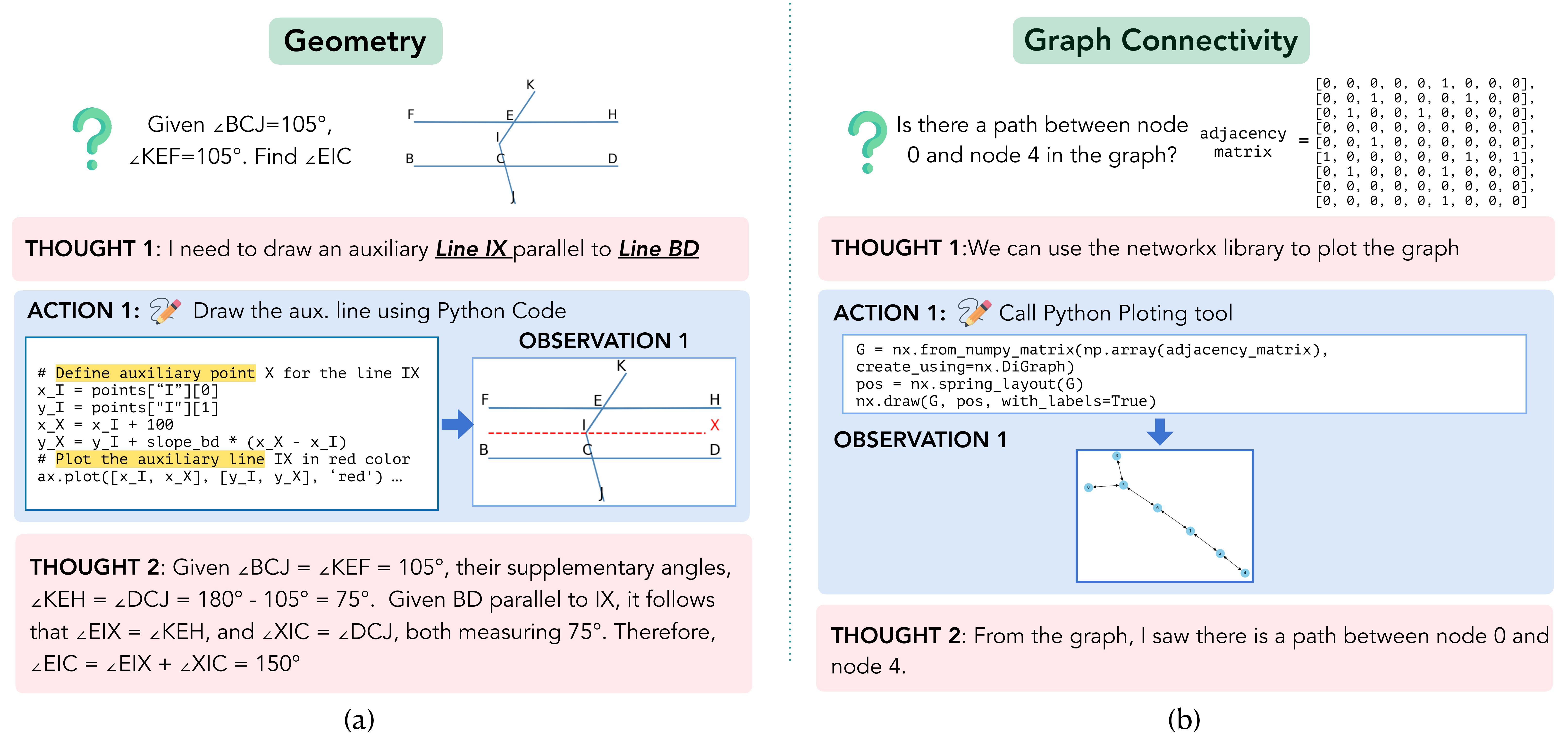}
    \caption{\textbf{Overview of \name}. Given a multimodal query, the \name agent generates a sketching plan to address the query (\textit{Thought}), and then synthesizes a program to create visual sketches (\textit{Action}). By analyzing the resulting sketches (\textit{Observation}), which serve as a visual representation of the reasoning process, the model generates a final response to the query. }
    \label{fig:method}
    \vspace{-1em}
\end{figure}

We introduce visual \name, a general framework that enables multimodal LMs to draw sketches as intermediate reasoning steps and to use these sketches to facilitate further reasoning. \autoref{fig:method} shows examples of how \name works. Given a multimodal query, \name agent generates a sketching plan to address the query (\textit{Thought}), and then synthesizes a program to create visual sketches (\textit{Action}). By analyzing the resulting sketches (\textit{Observation}), which serve as a visual representation of the reasoning process, the model generates a final response to the query. 

Our framework requires no finetuning or training. Multimodal LMs, out of the box, can be prompted to sketch using our framework.
Our implementation is based on the AutoGen~\cite{wu2023autogen} framework.
We give the overview of our \name framework in \S\ref{sec:method:react}, and delve deep into how it integrates sketching into the reasoning process in \S\ref{sec:method:sketch}. 
 
\subsection{Overview of \name}
\label{sec:method:react}

The \name agent solves tasks by engaging in an iterative interaction process with an environment. Given a multimodal query $q$ that includes both visual and textual components, the model generates a series of thoughts, actions, and observations to gather the information needed to answer the query.
At each time step $t$, the model performs three key steps:

\noindent{\textbf{Thought:}} The model analyzes the current context $c_t$, which includes the query, previous thoughts, actions, and observations, to generate a thought plan $p_t$ for the next action. For example, given the query $q$ -- ``\textit{find the $\angle EIC$}'' in \autoref{fig:method}a, the model's thought plan $p_1$ is to draw an auxiliary line $IX$ parallel to $BD$ serving  as a \textbf{\textit{visual sketch}} to help solve the problem.

\noindent{\textbf{Action:}} Based on the thought plan, the model executes an action $a_t$, which can manipulate both visual and textual content. In the geometry example, to realize the proposed thought of drawing the auxiliary line, the model generates Python code to modify the original geometry diagram. The generated code is then compiled and executed.

\noindent{\textbf{Observation:}} Based on the action $a_t$, \name's environment returns a new observation $o_{t+1}$, such as a new diagram with the auxiliary line drawn in the geometry example. The multimodal context is then updated to $c_{t+1} = (c_t, p_t, a_t, o_{t+1})$. 

The multi-turn interaction process continues until time step $T$, when the model determines that it has gathered enough information from the context $c_T$ to answer the query. At this point, it generates a special \textbf{Terminate} action and provides the answer.

Different from prior work \citep{yao2022react}, where LMs primarily generate and manipulate text-based observations and actions, \name enables the model to work with \textbf{multimodal observations $o_t$ and actions $a_t$, manipulating both visual and textual content.} This allows the model to plan and reason with the visual sketches it has drawn, enhancing its problem-solving capabilities.
% \name is implemented using the AutoGen~\cite{wu2023autogen} framework, which provides the necessary tools for the model to interact with a multimodal environment. 

% The multi-turn interaction process continues until time step $T$, when the model determines that it has gathered enough information from the context $c_T$ to answer the query. At this point, it generates a special \textbf{Terminate} action and provides the answer.
% In prior work, since most LMs only generate text, the observations $o_t$ are text-only, and actions $a_t$ only add or modify the text content. By contrast, \name \textbf{observations $o_t$ are multimodal, and its actions $a_t$ can manipulate both visual and textual content.} This enables LMs to plan and reason with the visual sketches it has drawn, enhancing its problem-solving capabilities.

% \autoref{fig:method} (a) illustrates an example of this process for a geometry problem. Given a visual input and a query $q$ such as "\textit{finding the $\angle EIC$}", the model generates a sketching plan (thought $p_1$) like "\textit{drawing an auxiliary line $IX$ parallel to line $BD$}". It then generates a program to draw the auxiliary line (action $a_1$) and receives a new diagram with the sketched auxiliary line (observation $o_1$). The multimodal context $c_t=(q, p_1, a_1, o_1)$ forms the basis for the model's next thought plan. In this case, the model can utilize the auxiliary line and introduced angles to solve the problem.

\subsection{Sketching via Code Generation}
\label{sec:method:sketch}
The core component of \name is sketching, which enables the LM to generate visual sketches by synthesizing programs that call different specialist vision models or Python plotting packages.

% \noindent\textbf{Program generation.} \weijia{move this similar to to later part} Similar to recent ViperGPT and VPD~\cite{gupta2023visual, surismenon2023vipergpt, hu2023visual}, we enable LMs to sketch through code generation. 
% We provide the LM with a detailed description of available tools that can generate multimodal content. An example prompt of description and prompt can be found in \S\ref{sec:appendix:prompt}.  We describe these modules later in \S\ref{sec:experiments_math} and \S\ref{sec:experiments_vision}. Specifically, there is a special \textit{display} function that allows the LM to ``visualize'' the image in the next observation $o_{t+1}$. Notice that the prompt does not contain the full implementation of these modules. Instead, it only provides Python function signatures and docstrings~\cite{hsieh2023tool}.  The LM generates Python code in a code block, using the provided tools, which when executed, generates new image and text outputs. 

\noindent{\textbf{Program Generation.}} Similar to recent works like ViperGPT and VPD~\cite{gupta2023visual, surismenon2023vipergpt, hu2023visual}, \name enables LMs to sketch through code generation. The LM is provided, through a prompt,
% \nascomment{, through a prompt,}
with a detailed description of the available tools that can generate multimodal content (an example prompt and description can be found in \S\ref{sec:appendix:prompt}). The prompt includes Python function signatures and docstrings~\cite{hsieh2023tool} for these modules, but does not contain their full implementation. The LM generates Python code in a code block, using the provided tools, which, when executed, generates new image and text outputs. A special \textit{display} function allows the LM to \textbf{visualize} the sketch image in the next observation $o_{t+1}$.

% The full prompt is in \S\ref{sec:appendix:prompt}.

\noindent{\textbf{Modules for sketching.}}
\name uses various tools to facilitate the sketching process, depending on the task at hand. For mathematical tasks, \name uses common Python packages like \texttt{matplotlib} and \texttt{networkx} for plotting (see \S\ref{sec:experiments_math}). For vision tasks, the LM leverages \textbf{specialist vision models} during the sketching process. These models include detection tools that draw bounding boxes on the image, as well as segmentation and marking tools (inspired by SoM~\cite{yang2023setofmark}) that draw colorful masks on the image and use numbers to label segments. We find these specialists possess useful perception skills for visual reasoning tasks, and \name is an effective way to combine them into a multimodal LM (see \S\ref{sec:experiments_vision:tools}).

% \subsection{Multimodal LMs for \name}
% As seen above, a multimodal LM that works well in \name should be good at multi-turn interaction, coding, and multimodal understanding. In all our experiments, we use  GPT-4-turbo-2024-04-09~\cite{gpt4} and GPT-4o-2024-05-13~\cite{gpt4}, since these models possess the above abilities. While we believe open-source LMs will evolve rapidly, we find that existing open-source multimodal LMs do not perform well on multi-turn interaction and coding. Nevertheless, we experiment with a simplified scenario in \S\ref{sec:analysis:open-source}, showing that open-source models using the oracle plan generated by GPT-4o also achieve great performance improvement.

% \nascomment{something I don't understand: what training/adaptation/finetuning needs to be done to get the LM to use this stuff?  or is it all in the prompt?  say more about how you make it work, after now have presented what it does ...} \ranjay{I added a sentence about this in response to Noah's comment at the beginning of this section.}
\section{Sketching to Solve Math Problems}
\label{sec:experiments_math}
% \weijia{Diagram is not very accurate. Better word? Our tasks cover graph, math, chess}

% \name is applicable to math-related tasks such as geometry questions, graph and  wehre the model can ... to use visual clues to to .... We selecting three different evaluation settings to showcase sketchpad abiltiy to helpful. the tasks we consider are: (1)  math, graph and chess 

In this section, we experiment with \name on four complex mathematical tasks
% in four different settings
: (1) geometry, (2) mathematical functions, (3) graph algorithms, and (4) game strategies. 
% Experiment setup including details of each task and employed sketching tools are discussed in \S \ref{sec:math-setup}. 
We demonstrate that incorporating sketching capabilities into LMs significantly improves their performance on these mathematical problems, setting new state-of-the-art results (\S \ref{sec:math-results}). 
% \weijia{results pending?}。 
% Finally, we show that open-source models can also improve upon sketches (\S\ref{sec:experiments_math:opensource}).

% This section details the experimental setup (\S \ref{sec:math-setup}). 
% \xingyu{We demonstrate that sketching on these math problems largely enhances multimodal LMs' mathematical reasoning abilities and set up new SOTAs on these tasks(\S \ref{sec:math-results}). Finally, we show that open-source models can also improve upon sketches(\S\ref{sec:experiments_math:opensource}). }

% \subsection{Experimental Setup}
%\label{sec:math-setup}
Details of our evaluation tasks and the tools employed for visual reasoning are as follows:

\noindent\textbf{Geometry Problems.}
Drawing auxiliary lines in geometry diagrams is often helpful for problem-solving. For example, in \autoref{fig:method} (a), when asked to find $\angle EIC$, the LM plans to draw an auxiliary line $IX$ parallel to $BD$, allowing it to use the properties of parallel lines to determine $\angle EIC$. To evaluate the effectiveness of \name, we use the problems from the Geometry3K dataset \citep{lu2021inter}.

To realize the line drawing process, \name takes a geometry diagram and its corresponding \texttt{matplotlib} code as input. The model then proposes and modifies the code to generate auxiliary lines, and executes the modified code to visualize the updated diagram with the added lines.
% \noindent\textbf{Geometry Problems}
% In geometry problems, it is often helpful to add auxiliary lines or segments to a diagram to facilitate problem-solving or concept proving. As shown in \autoref{fig:method} (a), given the query $q$ -- ``\textit{finding the $\angle EIC$}'' in \autoref{fig:method} (a), the model's thought plan $p_1$ is to draw an auxiliary line $IX$ parallel to $BD$ as it allows the model to use the properties of parallel lines and transversals to find angle relationships. Then we can easily get the  $\angle EIC$ by ....
% By equipping LMs with \name, they can draw these auxiliary lines to help solve the problem. We evaluate our model on a collection of geometry problems from the Geometry3K dataset \citep{lu2021inter}. %As shown in \autoref{fig:method}, 

% To generate the auxiliary line: Given a geometry diagram with its Python \texttt{matplotlib} code that can be used to generate an identical diagram, \name proposes modeifies Python code snippets and add the code for drawing thel ine and executes them to visualize the updated diagram with the auxiliary line included.

\noindent\textbf{Mathematical functions.}
Understanding mathematical functions is crucial for various applications in science, engineering, and economics. 
We focus on two tasks related to mathematical functions from the IsoBench datasets \citep{fu2024isobench}:
\begin{itemize}
    \item \textbf{\textit{Classifying parity}} aims to determine whether a function is even, odd, or neither. Even functions satisfy $f(-x) = f(x)$ for all $x$, while odd functions satisfy $f(-x) = -f(x)$. 

    \item \textbf{\textit{Identifying convexity/concavity}} aims to determine whether a function is convex or concave. 

    % involves determining whether line segments connecting points on the graph lie above/below or on the graph, resulting in a "bowl-shaped" (convex) or "dome-shaped" (concave) appearance.
\end{itemize}
Existing LMs can only analyze functions and attempt to prove their properties analytically.
\footnote{For humans, the analytical approach is the correct way to tackle these math tasks. However, we observe that LMs are not good at analytical reasoning in math. They make errors when deducing $f(-x)$ and derivatives.} 
However, \name enables them to visually sketch functions to solve problems more efficiently. For instance, to determine the convexity of the function in \autoref{fig:main}b, \name allows the model to plot the function using \texttt{matplotlib}, and visually inspect its overall shape. 

% making the parity and convexity/concavity properties visually apparent. This approach vividly illustrates that humans can solve the problem at a glance by looking at the plotted function, whereas it could take minutes to solve without the visual aid.
% By plotting the functions on the sketchpad, the parity and convexity/concavity properties become visually apparent, enabling models to make accurate predictions based on the geometric properties of the plotted functions. 

\noindent\textbf{Graph algorithms.}
Many real-world problems, such as those related to computer networks and transportation systems, can be formulated as graph problems. We evaluate \name on three graph problems from IsoBench \citep{fu2024isobench}:
\begin{itemize}
\item \textbf{\textit{Graph connectivity}} determines whether there exists a path between two vertices in a graph.
\item \textbf{\textit{Maximum flow}} aims to find the maximum amount of flow that can be sent through a network from a source vertex to a sink vertex, subject to capacity constraints on the edges.
\item \textbf{\textit{Graph isomorphism}} tests whether two graphs are structurally equivalent.
% i.e., whether there exists a bijective mapping between their vertex sets that preserves the edge relationships.
\end{itemize}
Given an adjacency matrix of a graph like in \autoref{fig:method}(b), \name can draw the actual graph structure, using using Python's \texttt{networkx} library, enabling direct visual reasoning about graph properties and relationships. 
%For example, to determine the graph connectivity of two nodes in \autoref{fig:method}, \name visualizes the graph from the adjacency matrix by using Python's \texttt{networkx} library to draw the graph. With the visual representation, the sketchpad helps identify connected components and visually trace paths between vertices, making it easier to assess connectivity.

% becomes a straightforward visual question once it is sketched, whereas the pure text version of the question can be more difficult to answer.
% To visualize the graph, our model uses Python's \texttt{networkx} library to draw the graph from the adjacency matrix. This visual representation enhances the model's ability to analyze graph properties, identify patterns, and solve graph-related problems more effectively.

% Graph problems have wide-ranging applications in various domains such as computer networks, transportation systems, and social networks. We explore three graph problems:

% Incorporating the sketchpad allows LMS to do visual representation and manipulation of graphs, enabling models to reason about their properties and relationships in a more intuitive manner. We evaluate our approach on the IsoBench datasets \citep{fu2024isobench} and showcase the effectiveness of \name in solving these problems. 

\noindent\textbf{Game strategies.}
Chess games can be represented in various formats, including visual board states and textual move notations. Given only the textual move notations, \name can draw the visual representations of the chess board to analyze positions and formulate strategies. We evaluate the performance of \name on the winnder identification task from the IsoBench datasets \citep{fu2024isobench} that aims to find the outcome of a chess game (win for White, win for Black, or draw) based on the final board state.
To create the graphical board, \name uses Python's \texttt{chess} library to draw the board from the Forsyth-Edwards Notation (FEN) of chess.
% \begin{itemize}
% \item \textbf{\textit{Winner identification}} aims to find the outcome of a chess game (win for White, win for Black, or draw) based on the final board state.
% % \item \textbf{\textit{Chess puzzles}} aims to identify the most advantageous first move given a specific chess position.
% \end{itemize}
% \weijia{remove chess puzzles}

\begin{table*}[]
\small
\centering
\vspace{-1em}
{\fontsize{8pt}{11pt}\selectfont
\begin{tabular}{l|ccccccc}
\toprule[1.2pt]
& \multicolumn{1}{c}{\textbf{Geometry}} & \multicolumn{3}{c}{\textbf{Graph}} & \multicolumn{2}{c}{\textbf{Math}} & \multicolumn{1}{c}{\textbf{Game}} \\
\cmidrule(lr){2-2} \cmidrule(lr){3-5} \cmidrule(lr){6-7} \cmidrule(lr){8-8}
Model & Geometry & Maxflow & Isomorphism & Connectivity & Convexity & Parity & Winner ID \\
\hline
\rowcolor{lightgray}
\multicolumn{8}{l}{\textit{Prior LLMs without visual inputs}} \\
Gemini-Pro & $\backslash$ &  15.6 & 47.7 & 50.0 & 87.9 & 48.2 & 8.1 \\
Claude 3 OPUS & $\backslash$ &  56.3 & 50.0 & 82.0 & 93.0 & 77.6 & 74.4 \\
Mixtral 8x7B \citep{jiang2023mistral} & $\backslash$ &  8.6 & 50.0 & 62.5 & 69.1 & 41.7 & 7.4 \\
LLaMA-2-70B \citep{touvron2023llama} & $\backslash$ &  18.0 & 50.0 & 50.0 & 74.2 & 33.3 & 12.4 \\
\hline
\rowcolor{lightgray}
\multicolumn{8}{l}{\textit{Latest multimodal LLMs + Visual Sketchpad}} \\
% Claude 3 OPUS \\
% + Sketchpad \\
% \\
% \hline
GPT-4 Turbo & 37.5 & 32.8 & 62.5 & 66.0 & 57.0 & 80.5 & 50.4 \\
+ Sketchpad & 45.8 & 96.8 & 97.6 & 97.6 &  77.3 & 71.5 & 64.2 \\
& \scoreg{+8.3} & \scoreg{+64.0} & \scoreg{+35.1} & \scoreg{+31.6} & \scoreg{+20.3} & \scorer{-9.0} & \scoreg{+13.8} \\
% + Sketchpad & 45.8 & 63.3 & 64.2 & 95.1 & \textbf{93.1} & \textbf{93.1} & 74.3 \\
% & \scoreg{+8.3} & \scoreg{+30.5} & \scoreg{+1.7} & \scoreg{+29.1} & \scoreg{+25.4} & \scoreg{+12.6} & \scoreg{+23.9} \\

\hline
% + Sketchpad & \textbf{66.7} & \textbf{66.3} & \textbf{65.3} & \textbf{98.1} & 90.1 & 88.1 & \textbf{81.2} \\
GPT-4o & 62.5 & 25.0 & 50.8 & 96.1 & 87.2 & 84.4 & 61.1 \\
+ Sketchpad & \textbf{66.7} & \textbf{66.3} & \textbf{65.3} & \textbf{98.4} & \textbf{94.9} & \textbf{94.7} & \textbf{64.6} \\
& \scoreg{+4.2} & \scoreg{+41.3} & \scoreg{+14.5} & \scoreg{+2.3} & \scoreg{+7.7} & \scoreg{+10.3} & \scoreg{+3.5} \\
% + Sketchpad & & \textbf{66.7}  & \textbf{66.3}  & \textbf{98.4} & \textbf{95.3} & 71.8 &  \\
 % & \scoreg{+4.2} & \scoreg{+41.3} & \scoreg{+14.5} & \scoreg{+2.0} & \scoreg{+2.9} & \scoreg{+3.7} & \scoreg{+20.1} \\
\bottomrule[1.2pt]
\end{tabular}
}
\vspace{-1ex}
\caption{Accuracy scores on geometry problems, graph algorithms, mathematical functions, and game. 
\textbf{\name yields large performance gains on most tasks and outperform all baselines.}}
\vspace{-2ex}
\label{tab:math}
\end{table*}

\subsection{Results}
\label{sec:math-results}

We evaluate the performance of \name on multimodal LMs with API access, including \texttt{gpt-4-turbo-2024-04-29} and \texttt{gpt-4o-2024-05-13}. We compare these results to baselines without the Visual Sketchpad and other notable closed-source models, such as Claude 3 and Gemini-Pro, as well as open-source models like Mistral \citep{jiang2023mistral} and LLaMA-2 70B \citep{touvron2023llama}. 
% The results are in \autoref{tab:math}.

\noindent\textbf{Main results.} As shown in \autoref{tab:math}, \name consistently improves base model performance across all tasks, with an average improvement of 11.2\% for \texttt{GPT-4o} and 23.4\% for \texttt{GPT-4 Turbo}. 
In particular, we observe large 
%\nascomment{here and in caption -- I suggest not using ``significant/significantly'' if you haven't done statistical significance tests.  better to say ``large gains'' or ``consistent gains''} 
gains on graph algorithms such as maximum flow and connectivity. For instance, \texttt{GPT-4o} with \name achieves an accuracy of 66.3\% on the maximum flow problem, improving over the base model by 41.3\%. 
Similarly, \name substantially improves the performance on mathematical functions, with \texttt{GPT-4 Turbo} achieving over 90\% accuracy and \texttt{GPT-4o} over 88\% accuracy on convexity and parity classification tasks. % \nascomment{4o is only 88.1\% on parity, though}. 
Furthermore, we observe gains (3\% $\sim$ 10\%) on game strategies, demonstrating that drawn game boards drawn can improve reasoning about game strategies.
Overall, these results highlight the effectiveness of \name in enhancing the reasoning capabilities of multimodal language models across diverse domains.

% \noindent\textbf{Qualitative Examples}
\vspace{-0.05in}
\section{Sketching to Solve Computer Vision Tasks}
\vspace{-0.05in}

% introduced the BLINK benchmark~\cite{fu2024blink}, which finds that many core visual perception abilities are still missing from existing multimodal LMs—even though many computer vision specialist models possess such abilities.

\label{sec:experiments_vision}
% In this section, we conduct experiments on complex visual reasoning tasks. 
In this section, we experiment with \name on complex visual reasoning tasks.
Recent work (BLINK)~\cite{fu2024blink}  finds that many core visual perception abilities are still missing from existing multimodal LMs—even though many computer vision specialist models possess such abilities.
Also, SoM~\cite{yang2023setofmark} shows that drawing segmentation masks on images unleashes the strong visual grounding ability of GPT-4V.
We generalize these ideas with \name, allowing LMs to use \textbf{specialist vision models} to sketch. Details of these modules are in \S\ref{sec:experiments_vision:tools}.
\name enhances multimodal LMs' visual reasoning abilities and establishes new SOTAs on all 7 tasks (\S\ref{sec:experiments_vision:results}). 
% Compared to visual prompting and other vision tool-use frameworks, \name is the only one that consistently improves performance across all tasks (\S\ref{sec:experiments_vision:compare}). 
% % We use the same set of modules for all experiments.

\noindent\textbf{Tasks.}
We experiment with a wide range of complex visual reasoning tasks:  (1) \textbf{$V^*$Bench}~\cite{wu2023vstar}. This benchmark contains questions about small items in an image. (2) \textbf{MMVP} benchmark from \textit{Eyes Wide Shut}~\cite{tong2024eyes}. This benchmark contains visual questions specially designed to reveal the visual shortcomings of CLIP-based multimodal LMs. (3) \textbf{BLINK}~\cite{fu2024blink}. This benchmark contains visual perception tasks that are easy for humans, but post significant challenge for multimodal LMs. Specifically, we experiment with relative depth, spatial reasoning, jigsaw puzzle, visual correspondence, and semantic correspondence tasks. More details of each task are in \S\ref{sec:appendix:dataset_stat}.

 \begin{figure}[h]
    \centering
    \includegraphics[width=0.95\linewidth]{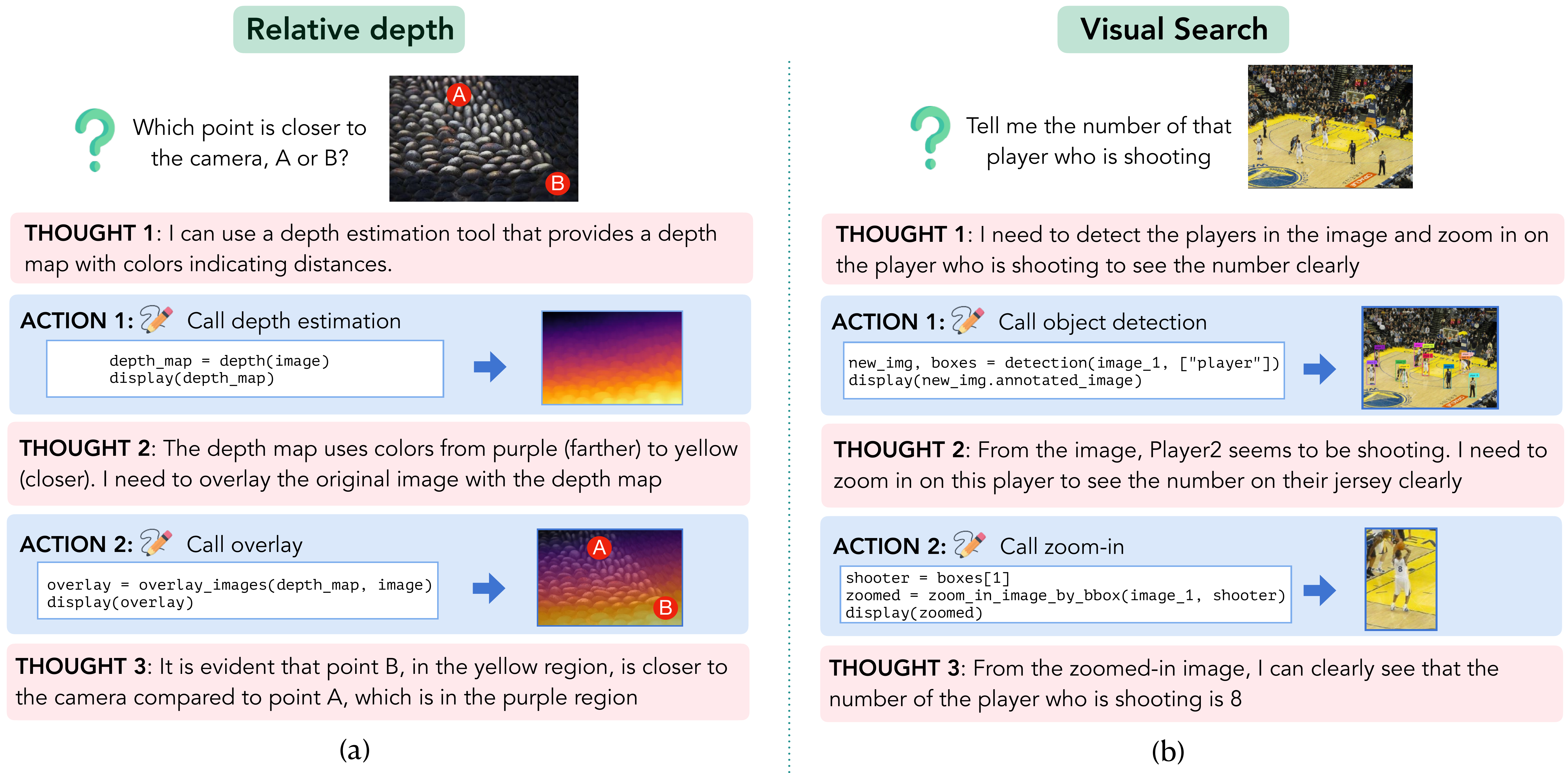}
    \caption{Examples of \name{} applied to vision tasks. The figure shows actual outputs generated by \name{}. By contrast, the baseline GPT-4o model cannot answer these questions correctly. Note that for demonstration purposes, the ``A'' and ``B'' marks in (a) are different from the actual images in the experiments.}
    \label{fig:vision_example}
    \vspace{-5mm}
\end{figure}

\vspace{-0.1in}

% \subsection{Vision specialist modules in \name}
\subsection{Vision Specialists as Sketching Tools in \name}
\label{sec:experiments_vision:tools}

LMs can use the following modules to sketch and manipulate images. We wrap these modules into Python functions that the LMs can call. Refer to \S\ref{sec:appendix:prompt} for the function definitions.

\noindent\textbf{Detection.} This module takes an image and a simple text query (e.g., ``cat'') as input. We run the Grounding-DINO~\cite{liu2023grounding} open-vocabulary objection detection model and plot the detected bounding boxes (together with a number label) on the image. It also returns the bounding box coordinates.

\noindent\textbf{Segmentation.} This module takes an image as input and returns an image with colorful segmentation masks on it. Each mask also has a number label. We follow the implementation of SoM~\cite{yang2023setofmark}. The underlying segmentation models are SegmentAnything~\cite{kirillov2023segment} and Semantic-SAM~\cite{li2023semantic}.

\noindent\textbf{Depth estimation.} This module takes an image as input and returns a depth map. The underlying model is DepthAnything~\cite{depthanything}.

\noindent\textbf{Visual search via sliding window.} This module mimics how humans search for small items on an image. It takes a text query as input and runs a sliding window over the image. The window size is 1/3 of the image size, and the step size is 2/9 of the image size (so an image will have $4 \times 4 = 16$ windows). It returns a sequence of image patches in which the query is detected.

\noindent\textbf{Other image manipulation modules.} Other modules include (1) \textbf{zoom-in and crop}, which takes an image and a bounding box as input and returns the image patch inside the box; (2) \textbf{Overlay images}, which takes two images and alpha values as input, and returns the overlayed image.

\vspace{-0.05in}
\subsection{Results}
\label{sec:experiments_vision:results}

We experiment with the same multimodal LMs as in \S\ref{sec:experiments_math} on complex visual reasoning tasks. We compare the performance with and without \name, as well as other notable multimodal LMs, including Gemini~\cite{team2023gemini}, Claude 3~\cite{claude}, and the open-source LLaVA 1.5~\cite{liu2023improvedllava}, LLaVA-NeXT~\cite{liu2024llavanext}.

\begin{table*}[h]
\small
\centering
{\fontsize{8pt}{11pt}\selectfont
\begin{tabular}{l|ccccccc}
\toprule[1.2pt]
Model & $V^*$Bench & MMVP & Depth & Spatial & Jigsaw & Vis. Corr. & Sem. Corr. \\
\hline
\rowcolor{lightgray}
\multicolumn{8}{l}{\textit{Prior multimodal LLMs}} \\
LLaVA-1.5-7B~\cite{liu2023improvedllava} & 48.7 & - & 52.4 & 61.5 & 11.3 & 25.6 & 23.0 \\
LLaVA-1.5-13B~\cite{liu2023improvedllava} & - & 24.7 & 53.2 & 67.8 & 58.0 & 29.1 & 32.4 \\
LLaVA-NeXT-34B~\cite{liu2024llavanext} & - & - & 67.7 & 74.8 & 54.7 & 30.8 & 23.7 \\
Claude 3 OPUS~\cite{claude} & - & - & 47.6 & 58.0 & 32.7 & 36.6 & 25.2 \\
Gemini-Pro~\cite{team2023gemini} & 48.2 & 40.7 & 40.3 & 74.8 & 57.3 & 42.4 & 26.6 \\
GPT-4V-preview~\cite{gpt4} & 55.0 & 38.7 & 59.7 & 72.7 & 70.0 & 33.7 & 28.8 \\
Previous state of the art & 75.4~\cite{wu2023vstar} & 49.3~\cite{gao2024sphinx} & 67.7~\cite{liu2024llavanext} & 76.2~\cite{2023internlm} & 70.0~\cite{gpt4} & 42.4~\cite{team2023gemini} & 33.1~\cite{wang2023cogvlm} \\
\hline
\rowcolor{lightgray}
\multicolumn{8}{l}{\textit{Latest multimodal LLMs + Visual Sketchpad}} \\
GPT-4 Turbo & 52.5 & 71.0 & 66.1 & 68.5 & 64.7 & 48.8 & 30.9 \\
+ Sketchpad & 71.0 & 73.3 & 68.5 & 80.4 & 68.5 & 52.3 & 42.4 \\
& \scoreg{+18.5} & \scoreg{+2.3} & \scoreg{+2.4} & \scoreg{+11.9} & \scoreg{+3.8} & \scoreg{+3.5} & \scoreg{+11.5} \\
\hline
GPT-4o & 66.0 & 85.3 & 71.8 & 72.0 & 64.0 & 73.3 & 48.6 \\
+ Sketchpad & \textbf{80.3} & \textbf{86.3} & \textbf{83.9} & \textbf{81.1} & \textbf{70.7} & \textbf{80.8} & \textbf{58.3} \\
& \scoreg{+14.3} & \scoreg{+1.0} & \scoreg{+12.1} & \scoreg{+9.1} & \scoreg{+6.7} & \scoreg{+7.5} & \scoreg{+9.7} \\
\bottomrule[1.2pt]
\end{tabular}
}
\vspace{-1ex}
\caption{
Accuracy on complex visual reasoning tasks. \textbf{\name enhances both GPT-4 Turbo and GPT-4o performance, establishing new SOTA performance levels on all the tasks.}
}
\vspace{-1em}
\label{tab:vision_main}
\end{table*}

\noindent\textbf{Main results.}
Table~\ref{tab:vision_main} shows the performance of our \name and baselines. \name consistently improves base model performance across all tasks. \texttt{GPT-4o} with \name sets the new state-of-the-art results on all tasks.
\name is particularly effective on $V^*$Bench, yielding 18.5\% accuracy improvement for \texttt{GPT-4 Turbo} and 14.3\% improvement for \texttt{GPT-4o}, surpassing the previous state of the art SEAL~\cite{wu2023vstar} which used a visual search model specifically trained for this task. On BLINK tasks, \name on average yields 6.6\% absolute accuracy gain for \texttt{GPT-4 Turbo} and 9.0\% gain for \texttt{GPT-4o}. Interestingly, despite the fact that all modules in \name work on a single image, the LMs also get substantial improvement on multi-image tasks, including jigsaw puzzles, visual correspondence, and semantic correspondence. Finally, \texttt{GPT-4o}, the LM with stronger multimodal ability than \texttt{GPT-4 Turbo}, benefits more from \name. For example, on the relative depth task, \texttt{GPT-4o} gets 12.1\% accuracy improvement, while \texttt{GPT-4 Turbo} only gets 2.4\%, showing that \texttt{GPT-4o} is better at understanding the depth map \name generated. Overall, our experiments show that \name is an effective way to improve multimodal LMs' performance on visual reasoning tasks.

\noindent\textbf{How many times is each vision specialist used?}
We count the number of times each vision specialist is used in each task, as shown in Figure~\ref{fig:tool-use frequency}. Here we choose the four tasks that achieve the largest improvement: $V^*$Bench, relative depth, spatial reasoning, and semantic correspondence. 
%The $y$-axis is the frequency with which each function is used. 
We observe that  (1) \textbf{the use of vision specialist is task-dependent, and the two LMs analyzed utilize similar tools.} For example, for $V^*$, which needs to locate small objects, the LMs mainly use detection, sliding window search, and zoom-in, similar to how people would search. For the relative depth task, both models rely on depth estimation. For spatial reasoning, the LMs use detection and segmentation to facilitate visual reasoning. 
(2) \textbf{GPT-4o likes to use more tools.} 
\texttt{GPT-4o} uses the vision specialists more often than \texttt{GPT-4 Turbo.} Also, the two LMs behave differently for the semantic correspondence tasks. \texttt{GPT-4o} uses the segmentation module for $40\%$ of the task instances, while \texttt{GPT-4 Turbo} uses the detection module for less than $20\%$ of times, and rarely uses the segmentation module. This difference may explain the performance gap between the two LMs (58.3\% v.s. 42.4\%) on this task.

\begin{table*}[h]
\small
\centering

\begin{minipage}{0.48\textwidth}
\centering
        \includegraphics[width=\linewidth]{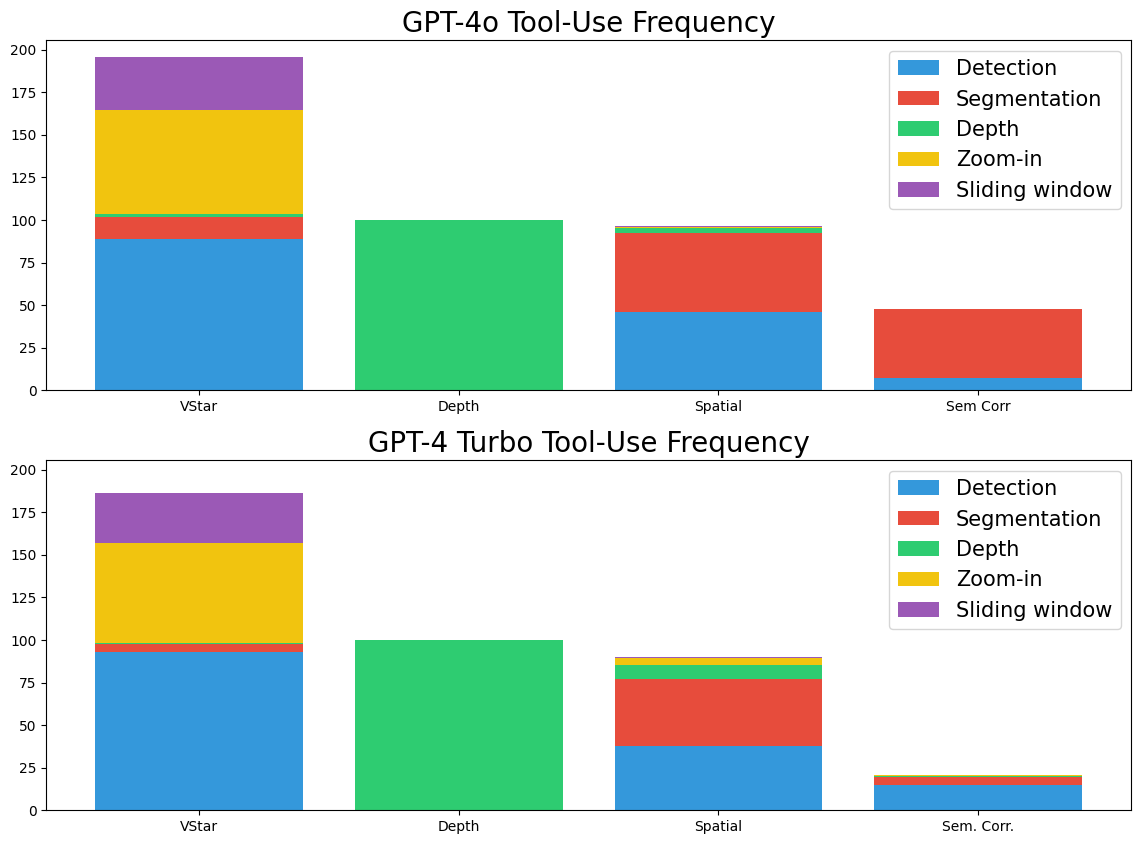} 
        \captionof{figure}{Percentage of times GPT-4o and GPT-4 Turbo use a visual module in \name when solving $V^*$Bench, relative depth, spatial reasoning, and semantic correspondence tasks.}
        \label{fig:tool-use frequency}
\end{minipage}
\hfill
\begin{minipage}{0.47\textwidth}
{\fontsize{8pt}{10pt}\selectfont
\begin{tabular}{l|cccc}
\toprule[1.2pt]
Model & $V^*$ & MMVP & Depth & Spatial \\
\midrule
\rowcolor{lightgray}
GPT-4 Turbo & 52.5 & 71.0 & 66.1 & 68.5 \\
$\ \ $SoM & 42.0 & 60.7 & 58.9 & 78.3 \\
$\ \ $SoM + orig.& 51.3 & \textbf{74.3} & 66.9 & 79.7 \\
$\ \ $Visprog & 33.2 & 16.3 & 67.8 & 53.8\\
$\ \ $Sketchpad & \textbf{71.0} & 73.3 & \textbf{68.5} & \textbf{80.4} \\
\midrule
\rowcolor{lightgray}
GPT-4o & 66.0 & 85.3 & 71.8 & 72.0 \\
$\ \ $SoM & 49.0 & 70.7 & 62.9 & \textbf{83.2} \\
$\ \ $SoM + orig. & 68.1 & 84.0 & 75.0 & 82.5 \\
$\ \ $Visprog & 32.4 & 17.3 & 46.8 & 37.8\\
$\ \ $Sketchpad & \textbf{80.3} & \textbf{86.3} & \textbf{83.9} & 81.1 \\
\bottomrule[1.2pt]
\end{tabular}
}
\vspace{-1ex}
\caption{
Comparison with other augmentation frameworks for multimodal LMs on single-image tasks.  For fair comparison, we modify the original Visprog~\cite{gupta2023visual} framework by replacing the LM and VQA components with the corresponding GPT-4 model.
\vspace{-2ex}
}
\label{tab:vision_compare}
\end{minipage}
\end{table*}

%\subsection{Comparison with visual prompting and tool-use frameworks}
% \label{sec:experiments_vision:compare}

\noindent\textbf{Comparison with visual prompting and tool-use frameworks.}
In Table~\ref{tab:vision_compare}, we compare \name with the visual prompting framework \textbf{SoM}~\cite{yang2023setofmark} and the LLM tool-use framework \textbf{Visprog}~\cite{gupta2023visual}. Details of these methods can be found in \S\ref{sec:related_work}.
% \noindent\textbf{Visual prompting and tool-use baselines.}
% We compare \name with other LM augmentation frameworks that use vision specialists. 
% The first is \textbf{SoM}~\cite{yang2023setofmark}, a visual prompting method that employs segmentation models to draw masks on images. GPT-4V shows enhanced visual grounding ability when taking these augmented images as inputs. 
% Compared with SoM, \name introduces more vision tools beyond segmentation and allows LM to choose which tools to use.
% The other baseline is \textbf{Visprog}~\cite{gupta2023visual},  a framework that uses LM-generated codes to compose vision specialists into subroutines to solve visual reasoning tasks.
% In Visprog, the vision specialists stick to the plan defined in the codes during execution. In contrast, \name can adjust its plan after observing intermediate visual artifacts, which is more effective and robust for multimodal reasoning. 
% %\ranjay{These baselines are already explained in related work. I would cut this paragraph and replace this subsection with just a single paragraph.}
For a fair comparison, we make the following adaptations: (1) we find that prompting LMs with SoM images can hurt performance, likely because the visual prompts confuse the model. To make a stronger baseline, we prompt the LM with both the original image and the SoM image (full prompt in \S\ref{sec:appendix:prompt}), which we refer as ``SoM + orig.'' (2) We replace the LM and VQA modules in Visprog with the corresponding GPT-4 model. (3) Since baseline methods are developed on single-image tasks, we compare \name on such tasks. From Table~\ref{tab:vision_compare}, we can see that \textbf{\name is the only framework that yields consistent improvement on all tasks.} SoM can boost spatial reasoning ability, as the authors reported. However, it can hurt the performance on other tasks, even in the ``SoM + orig.'' setting. Visprog performs worse than the base LM on all the tasks. As prior work~\cite{khandelwal2023analyzing, hu2023visual} suggests, one possible reason is that the vision modules themselves have errors, and the error propagates when the modules are composed by a program.

\vspace{-0.08in}
\section{Analysis and Discussion}
\label{sec:analysis}
\vspace{-0.05in}

\noindent\textbf{Why does \name work?} 
First, \textbf{vision is a versatile and informational interface that complements language}. Dense information like depth and segmentation cannot be described easily through language~\cite{fu2024blink}. In a broader perspective, humans have developed many visualization techniques that are direct, efficient, and informational. \name provides LMs the opportunity to use them.
Second, in \name, multimodal LMs can \textbf{plan and reason based on the intermediate visual artifacts} they created. In contrast, in prior modular vision work~\cite{gupta2023visual, surismenon2023vipergpt, yang2023setofmark}, multimodal modules follow a predefined plan by either humans or code. \name is much more flexible and robust to errors. For example, suppose object detection makes an error. The LM can (in principle) find the error by viewing the bounding boxes, and change its following plans, but prior methods cannot. 
Third, as discussed next, \textbf{the plans of multimodal LMs are similar to human plans}, and therefore likely benefit from the fact that the underlying LMs have seen  data with similar reasoning patterns. 
%suggesting that there might be plenty of ``\name-style visual chain-of-thought'' in multimodal pre-training data. \name unleashes models' ability to perform such complex multimodal reasoning.

\noindent\textbf{Do LMs have the same plans as humans?} We conduct a human study on all geometry problems and 10 problems on each vision task. On geometry, humans draw the same auxiliary line as GPT-4o 80\% of the time. On vision, we show 2 human subjects the full plan of GPT-4o, which they rate is valid in 92.8\% of instances. Most errors are caused by failures in the vision specialists (e.g., fail to detect an object) and mistakes in simple visual question answering, rather than planning.

% \nascomment{fix table 4 so all numbers use the same number of significant digits (one is enough)}

\begin{table*}[h]
\small
\centering
% \vspace{-1em}
% {\fontsize{8pt}{11pt}
\selectfont
\begin{tabular}{l|cccc}
\toprule[1.2pt]
Model & Geometry & Maxflow & Convexity & Winner ID \\
\hline
\rowcolor{lightgray}
LLaVA-NeXT-13B & 11.1 & 7.8 &50.39 & 5.8  \\
+ oracle Sketchpad & 22.2 & 10.2 & 50.0 & 36.7 \\
\hline
\rowcolor{lightgray}
LLaVA-NeXT-34B  & 26.1 & 0.8 &  81.6 & 49.0 \\
+ oracle Sketchpad & 28.3 & 14.1 & 87.1 & 49.4 \\
\bottomrule[1.2pt]
\end{tabular}
% }
% \vspace{-1ex}
\caption{Open-source LLaVA models' performance on math tasks. The oracle Sketchpad  uses the visual artifact generated in the last action of GPT-4o + \name as inputs.}
\vspace{-1ex}
\label{tab:open-source}
\end{table*}

\noindent\textbf{Experiments on open-source models. } Can sketches like diagrams, plots, and auxiliary lines facilitate existing open-source multimodal LMs? To answer this question, we conduct the experiments in Table~\ref{tab:open-source}. We use the visual artifacts generated in the last action of GPT-4o + \name experiment as the image input for open-source LLaVA-NEXT models~\cite{liu2024llavanext}. We can see that this oracle \name brings consistent improvement to math tasks and boosts mathematical reasoning.

% \noindent\textbf{Broader Impact.} Not only for machines, but also for humans
%\vspace{-1em}
\vspace{-0.05in}
\section{Conclusion}
\label{sec:conclusion}
\vspace{-0.05in}
We present Visual \name, a framework that provides multimodal LMs with the tools necessary to generate intermediate sketches to reason over tasks. For complex mathematical reasoning tasks, \name yields large performance gains, by visualizing auxiliary lines, math functions, graphs, and games during reasoning. For visual reasoning tasks, we add vision specialists to \name. The LM can call these specialists during reasoning, observing the visualization of these specialists' predictions (e.g., bounding boxes from the object detection model; masks from the segmentation model), and then conduct further planning and reasoning. Experiments show that \name  enhances the LMs' performance across all tasks, and sets new state-of-the-art results. Ultimately, \name represents a step toward endowing LMs with more human-like multimodal intelligence, leveraging the complementary strengths of language and vision to tackle increasingly complex reasoning challenges.

\noindent\textbf{Limitations and future directions.} First, \name requires more computing resources than directly outputting language tokens. 
% \nascomment{drop this sentence; it's subjective and depends on the user:} However, the performance gain makes up for the cost. \nascomment{this sentence feels unconnected from the rest:} Since existing models' multimodal reasoning abilities are far from satisfying~\cite{fu2024blink, tong2024eyes}, as multimodal LMs are deployed in more areas (e.g., robotics, education), more complex multimodal tasks will emerge and present new research opportunities. 
We discuss more about computing costs in~\ref{sec:appendix:cost}. Second, this work focuses on existing off-the-shelf LMs. Future work may explore the training side of \name. For example, recent multimodal models like Unified-IO 2~\cite{lu2023unified} and Chameleon~\cite{Team2024ChameleonME} are natively multimodal and can output both text and images. \name may emerge as a new paradigm for instruction tuning these models. Finally, \name can be applied in more areas. For example, for robotics, we can apply \name to search for small things in a crowded space, highlight the object of interest, and zoom the camera for a better view or use depth estimation to help navigation.

\clearpage
\bibliographystyle{plain}
\bibliography{main}

\newpage
\appendix
\appendixpage
\startcontents[sections]
\printcontents[sections]{l}{1}{\setcounter{tocdepth}{2}}
\newpage

\section{Acknowledgement}
This work is in part supported by gifts from Allen Institute for Artificial Intelligence, Meta, Apple, and Amazon Science. We also greatly appreciate the help from Zixian Ma, Wei-Chiu Ma, Benlin Liu, Jieyu Zhang, UW NLP, AllenNLP, AI2 PRIOR, and UW RAIVN for their help.

\section{More Qualitative Examples}
\label{sec:appendix:example}

Here we provide more details of how \texttt{GPT-4o} and \name solve complex mathematical and visual reasoning tasks.
\begin{figure}[h]
    \includegraphics[width=1.01\linewidth]{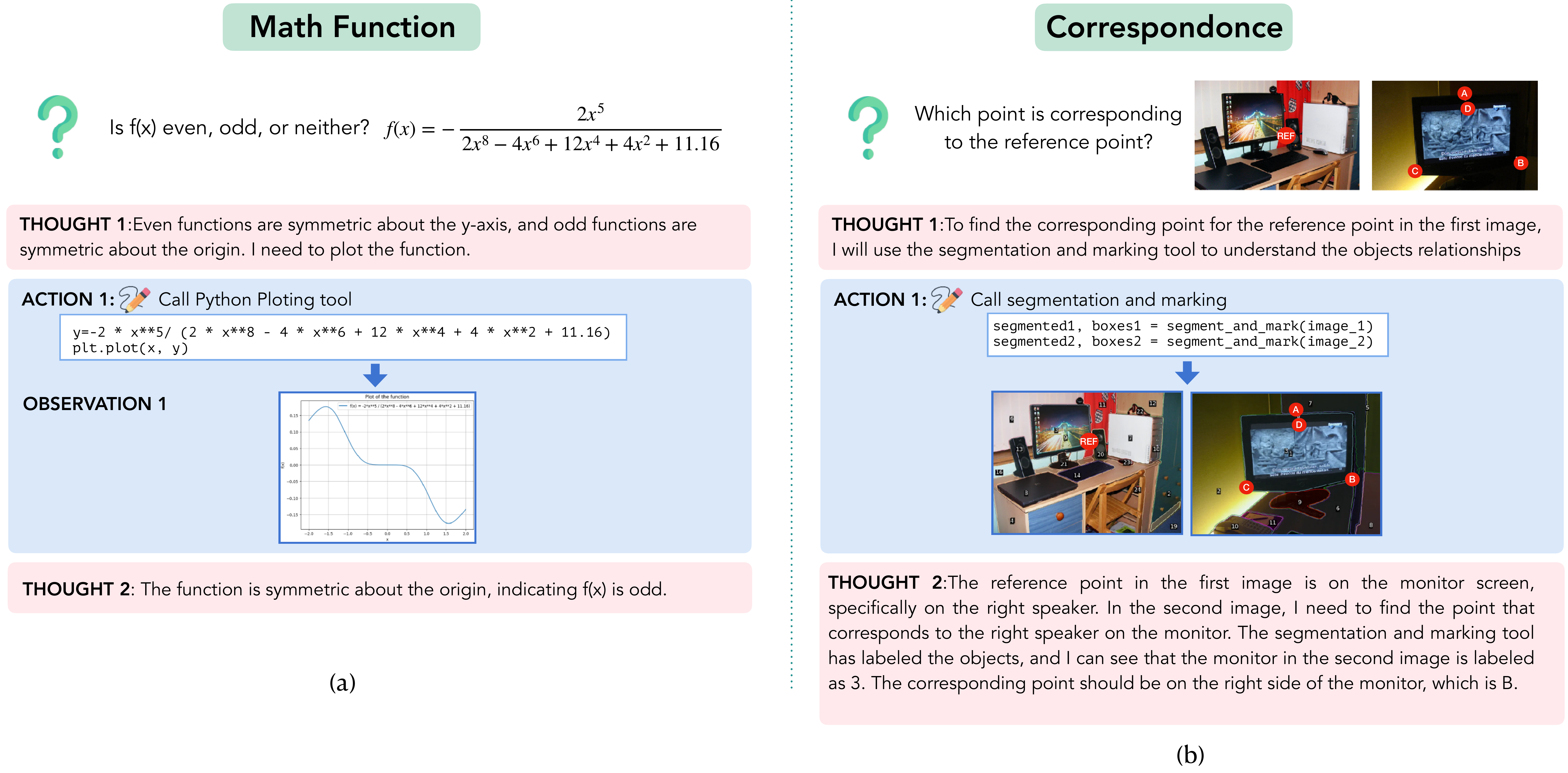}
    \caption{\textbf{Qualitative examples} of how \texttt{GPT-4o} and \name solve (a) complex mathematical and (b) visual reasoning tasks.}
    \label{fig:qual}
\end{figure}

\clearpage

\section{Prompts}
\label{sec:appendix:prompt}

\textbf{Prompts for computer vision tasks.} Here we provide the running log of GPT-4o + \name tackling a visual search problem from $V^*$Bench~\cite{wu2023vstar}. We use the same prompt template for all computer vision problems in this work. For visualization purposes, we present the prompts that contain codes differently. In our implementation, they are all text inputs to the LLM.
% \lstinputlisting[style=mystle]{figures/vision_input.py}
% Notice that despite the object detection model is making mistakes, GPT-4o can overcome these mistakes and get the correct answer.

\begin{figure}[h]
\noindent\makebox[\textwidth][c]{
\framebox{%
\begin{minipage}{1.0\linewidth}
\usefont{T1}{ppl}{m}{n} % Palatino font
\small
% \color{MidnightBlue} % Using a deep blue color for the text
\textbf{SYSTEM PROMPT}

You are a helpful multimodal AI assistant. [MORE INSTRUCTIONS ...]
$$\ $$
For each turn, you should first do a "THOUGHT", based on the images and text you see.
If you think you get the answer to the intial user request, you can reply with "ANSWER: <your answer>" and ends with "TERMINATE".
\end{minipage}
}
}
\end{figure}

\begin{figure}[h]
\noindent\makebox[\textwidth][c]{
\framebox{%
\begin{minipage}{1.0\linewidth}
\usefont{T1}{ppl}{m}{n} % Palatino font
\small
% \color{MidnightBlue} % Using a deep blue color for the text
\textbf{Initial Prompt + Request}
\end{minipage}
}
}
\end{figure}

\begin{lstlisting}[language=Python]
Here are some tools that can help you. All are python codes. They are in tools.py and will be imported for you.
The images has their own coordinate system. The upper left corner of the image is the origin (0, 0). All coordinates are normalized, i.e., the range is [0, 1].
All bounding boxes are in the format of [x, y, w, h], which is a python list. x is the horizontal coordinate of the upper-left corner of the box, y is the vertical coordinate of that corner, w is the box width, and h is the box height.
Notice that you, as an AI assistant, is not good at locating things and describe them with coordinate. You can use tools to generate bounding boxes.
You are also not good at answering questions about small visual details in the image. You can use tools to zoom in on the image to see the details. Below are the tools in tools.py:
```python
class AnnotatedImage:
    # A class to represent an annotated image. It contains the annotated image and the original image.
    
    def __init__(self, annotated_image: Image.Image, original_image: Image.Image=None):
        self.annotated_image = annotated_image
        self.original_image = original_image
        
def detection(image, objects):
    """Object detection using Grounding DINO model. It returns the annotated image and the bounding boxes of the detected objects.
    The text can be simple noun, or simple phrase (e.g., 'bus', 'red car'). Cannot be too hard or the model will break.
    The detector is not perfect, it may wrongly detect objects or miss some objects.
    Also, notice that the bounding box label might be out of the image boundary.
    You should use the output as a reference, not as a ground truth.
    When answering questions about the image, you should double-check the detected objects.

    Args:
        image (PIL.Image.Image): the input image
        objects (List[str]): a list of objects to detect. Each object should be a simple noun or a simple phrase. Should not be hard or abstract concepts like "text" or "number".

    Returns:
        output_image (AnnotatedImage): the original image, annotated with bounding boxes. Each box is labeled with the detected object, and an index.
        processed boxes (List): listthe bounding boxes of the detected objects
    
    Example:
        image = Image.open("sample_img.jpg")
        output_image, boxes = detection(image, ["bus"])
        display(output_image.annotated_image)
        print(boxes) # [[0.24, 0.21, 0.3, 0.4], [0.6, 0.3, 0.2, 0.3]]
        # you need to double-check the detected objects. Some objects may be missed or wrongly detected.
    """
    
def sliding_window_detection(image, objects):
    """Use this when you are searching for objects in the image, but the objects are not detected by the object detection model.
    In that case, the most common reason is that the object is too small such that both the vision-language model and the object detection model fail to detect it.
    This function tries to detect the object by sliding window search.
    With the help of the detection model, it tries to detect the object in the zoomed-in patches.
    The function returns a list of annotated images that may contain at leas one of the objects, annotated with bounding boxes.
    It also returns a list of a list of bounding boxes of the detected objects.

    Args:
        image (PIL.Image.Image): the input image
        objects (List[str]): a list of objects to detect. Each object should be a simple noun or a simple phrase. Should not be hard or abstract concepts like "text" or "number".
        
    Returns:
        possible_patches (List[AnnotatedImage]): a list of annotated zoomed-in images that may contain the object, annotated with bounding boxes.
        possible_boxes (List[List[List[Float]]]): For each image in possible_patches, a list of bounding boxes of the detected objects. 
            The coordinates are w.r.t. each zoomed-in image. The order of the boxes is the same as the order of the images in possible_patches.
            
    Example:
        image = Image.open("sample_img.jpg")
        possible_patches, possible_boxes = search_object_and_zoom(image, ["bird", "sign"])
        for i, patch in enumerate(possible_patches):
            print(f"Patch {i}:")
            display(patch.annotated_image)
        
        # print the bounding boxes of the detected objects in the first patch
        print(possible_boxes[0]) # [[0.24, 0.21, 0.3, 0.4], [0.6, 0.3, 0.2, 0.3]]
    """
    
def segment_and_mark(image, anno_mode:list = ['Mask', 'Mark']):
    """Use a segmentation model to segment the image, and add colorful masks on the segmented objects. Each segment is also labeled with a number.
    The annotated image is returned along with the bounding boxes of the segmented objects.
    This tool may help you to better reason about the relationship between objects, which can be useful for spatial reasoning etc.
    DO NOT use this tool to search or detect an object. It is likely the object is small and segmentaiton does not help.
    Segmentation and marking can also be helpful for 3D and video reasoning. For example, helping you to see more clearly and analyzes the relationship between different frames of a video.
    
    Args:
        image (PIL.Image.Image): the input image
        anno_mode (list, optional): What annotation is added on the input image. Mask is the colorful masks. And mark is the number labels. Defaults to ['Mask', 'Mark'].
        
    Returns:
        output_image (AnnotatedImage): the original image annotated with colorful masks and number labels. Each mask is labeled with a number. The number label starts at 1.
        bboxes (List): listthe bounding boxes of the masks.The order of the boxes is the same as the order of the number labels.
        
    Example:
        User request: I want to find a seat close to windows, where should I sit?
        Code:
        ```python
        image = Image.open("sample_img.jpg")
        output_image, bboxes = segment_and_mark(image)
        display(output_image.annotated_image)
        ```
        Model reply: You can sit on the chair numbered as 5, which is close to the window.
        User: Give me the bounding box of that chair.
        Code:
        ```python
        print(bboxes[4]) # [0.24, 0.21, 0.3, 0.4]
        ```
        Model reply: The bounding box of the chair numbered as 5 is [0.24, 0.21, 0.3, 0.4].
    """
    
def depth(image):
    """Depth estimation using DepthAnything model. It returns the depth map of the input image. 
    A colormap is used to represent the depth. It uses Inferno colormap. The closer the object, the warmer the color.
    This tool may help you to better reason about the spatial relationship, like which object is closer to the camera.

    Args:
        image (PIL.Image.Image): the input image

    Returns:
        output_image (PIL.Image.Image): the depth map of the input image
        
    Example:
        image = Image.open("sample_img.jpg")
        output_image = depth(image)
        display(output_image)
    """
    
def zoom_in_image_by_bbox(image, box, padding=0.05):
    """A simple wrapper function to crop the image based on the bounding box.
    When you want to answer question about visual details in a bounding box annotated by the detection tool, you would like to zoom in on the object using this function.

    Args:
        image (PIL.Image.Image): the input image
        box (List[float]): the bounding box in the format of [x, y, w, h]
        padding (float, optional): The padding for the image crop, outside of the bounding box. Defaults to 0.1. The zoom factor cannot be too small. Minimum is 0.05

    Returns:
        cropped_img (PIL.Image.Image): the cropped image
        
    Example:
        image = Image.open("sample_img.jpg")
        annotated_img, boxes = detection(image, "bus")
        cropped_img = zoom_in_image_by_bbox(image, boxes[0], padding=0.05)
        display(cropped_img)
    """
    
def overlay_images(background_img, overlay_img, alpha=0.3, bounding_box=[0, 0, 1, 1]):
    """
    Overlay an image onto another image with transparency.
    This is particularly useful visualizing heatmap while preserving some info from the original image.
    For example, you can overlay a segmented image on a heatmap to better understand the spatial relationship between objects.
    It will also help seeing the labels, circles on the original image that may not be visible on the heatmap.

    Args:
    background_img_pil (PIL.Image.Image): The background image in PIL format.
    overlay_img_pil (PIL.Image.Image): The image to overlay in PIL format.
    alpha (float): Transparency of the overlay image.
    bounding_box (List[float]): The bounding box of the overlay image. The format is [x, y, w, h]. The coordinates are normalized to the background image. Defaults to [0, 0, 1, 1].

    Returns:
    PIL.Image.Image: The resulting image after overlay, in PIL format.
    s
    Example:
        image = Image.open('original.jpg')
        depth_map = depth(image)
        overlayed_image = overlay_images(depth_map, image, alpha=0.3)
        display(overlayed_image)
    """
```
# GOAL #: Based on the above tools, I want you to reason about how to solve the # USER REQUEST # and generate the actions step by step (each action is a python jupyter notebook code block) to solve the request.
You may need to use the tools above to process the images and make decisions based on the visual outputs of the previous code blocks.
Your visual ability is not perfect, so you should use these tools to assist you in reasoning about the images.

# [six in-context examples here]

# USER REQUEST #: 
\end{lstlisting}

\begin{figure}[h]
\centering
    \includegraphics[width=0.8\linewidth]{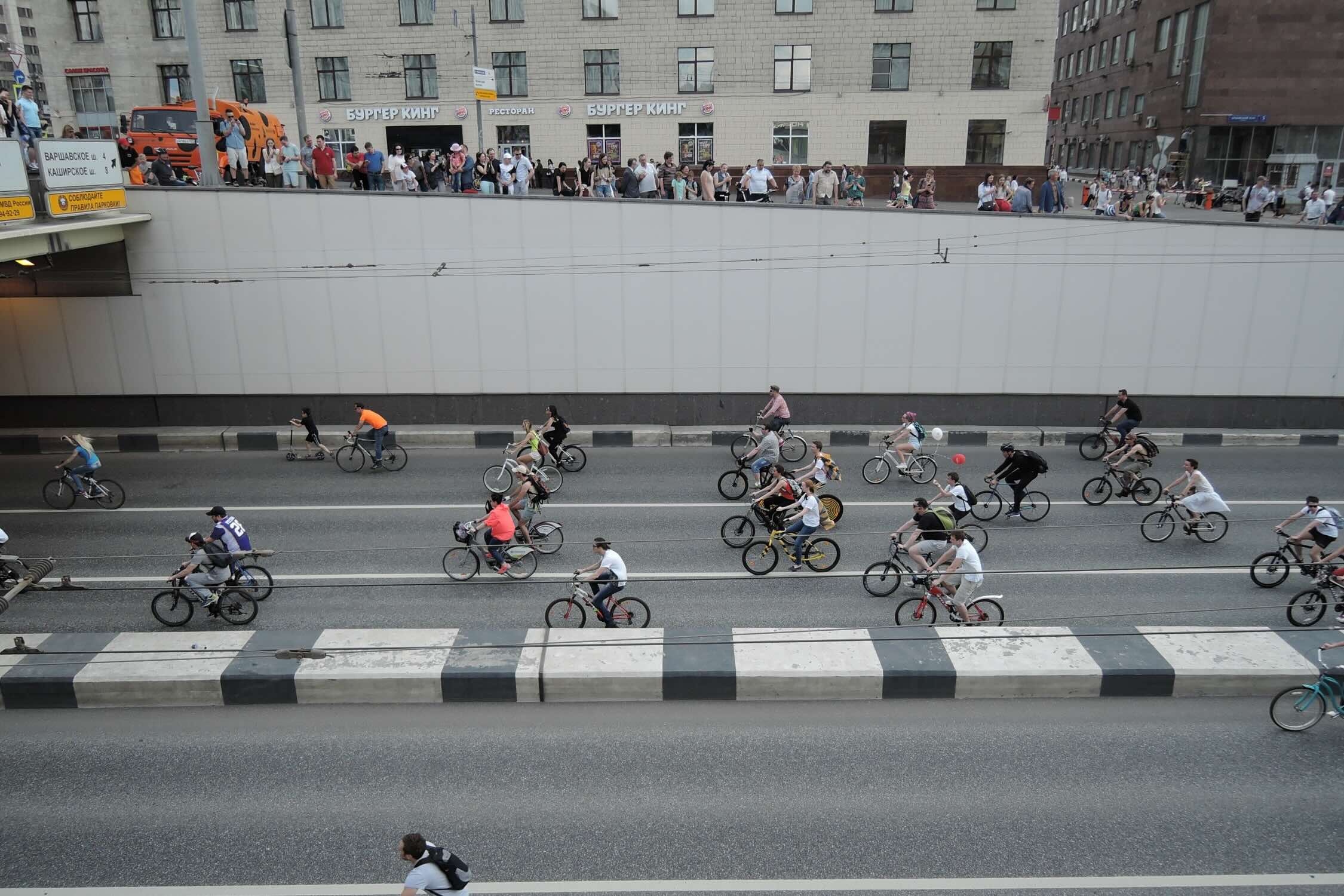}
\end{figure}

\begin{lstlisting}[language=Python]
Is the red balloon above or below the white balloon? Options: (A) The red balloon is below the white balloon. (B) The red balloon is above the white balloon.

# USER IMAGE stored in image_1 as PIL image.
Now please generate only THOUGHT 0 and ACTION 0 in RESULT. If no action needed, also reply with ANSWER: <your answer> and ends with TERMINATE in the RESULT:

# RESULT #:
\end{lstlisting}

\begin{figure}[h]
\noindent\makebox[\textwidth][c]{
\framebox{%
\begin{minipage}{1.0\linewidth}
\usefont{T1}{ppl}{m}{n} % Palatino font
\small
% \color{MidnightBlue} % Using a deep blue color for the text
\textbf{Thought + Action 0}
\end{minipage}
}
}
\end{figure}

\begin{lstlisting}[language=Python]
THOUGHT 0: To determine the relative positions of the red and white balloons, I need to detect the balloons in the image. I will use the detection tool to find the balloons.
ACTION 0:
```python
image_1_balloon_detection, image_1_balloon_boxes = detection(image_1, ["red balloon", "white balloon"])
display(image_1_balloon_detection.annotated_image)
```
\end{lstlisting}

\clearpage

\begin{figure}[h]
\noindent\makebox[\textwidth][c]{
\framebox{%
\begin{minipage}{1.0\linewidth}
\usefont{T1}{ppl}{m}{n} % Palatino font
\small
% \color{MidnightBlue} % Using a deep blue color for the text
\textbf{Observation 0}
\end{minipage}
}
}
$\ \ \\$
Execution success. The output is as follows. 

    \includegraphics[width=0.8\linewidth]{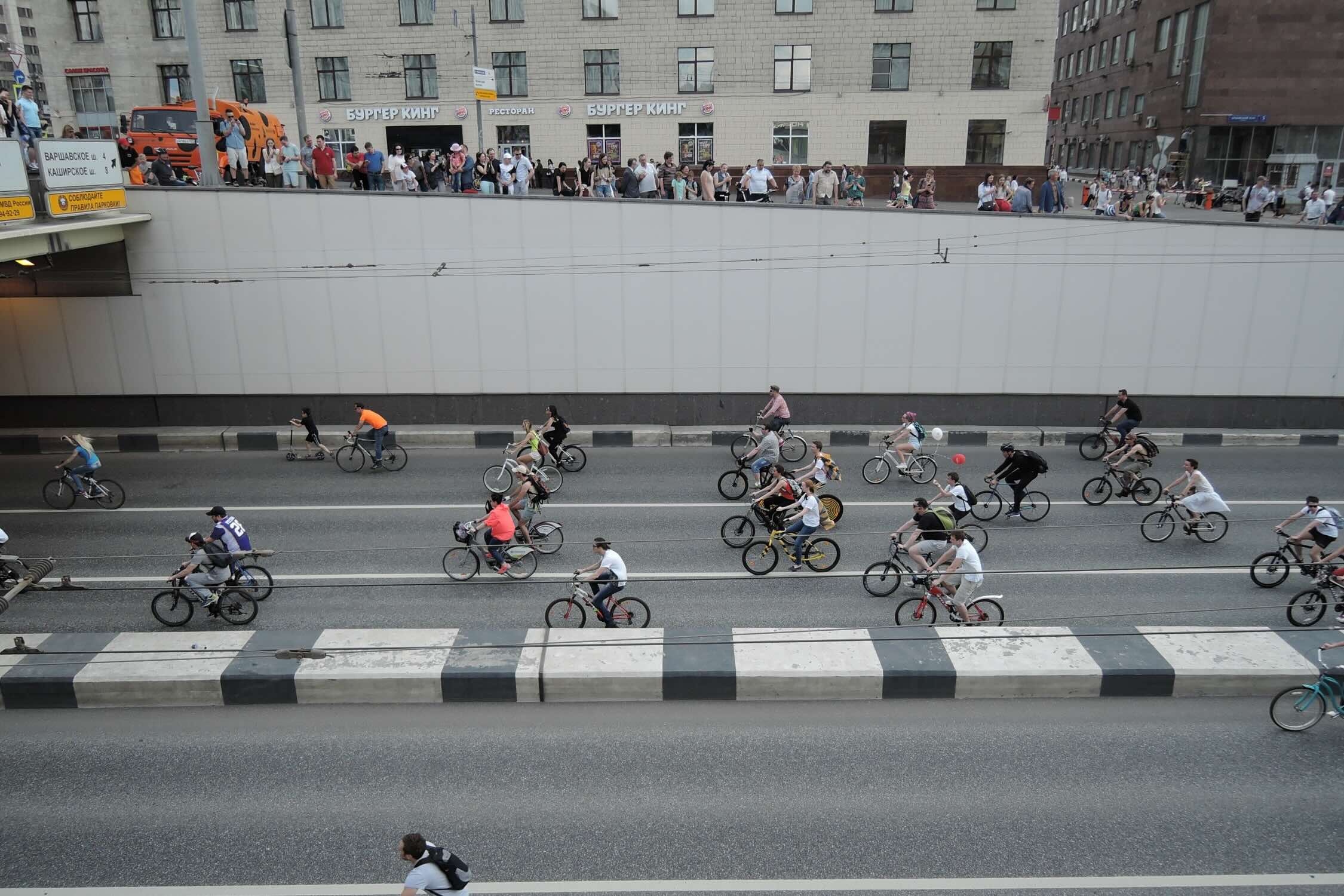}
    \noindent\makebox[\textwidth][c]{

\begin{minipage}{1.0\linewidth}
\usefont{T1}{ppl}{m}{n} % Palatino font
\small
% \color{MidnightBlue} % Using a deep blue color for the text
Please generate the next THOUGHT and ACTION. If you can get the answer, please also reply with ANSWER: <your answer> and ends with TERMINATE.

\end{minipage}
}

\vspace{0.3in}
\noindent\makebox[\textwidth][c]{
\framebox{%
\begin{minipage}{1.0\linewidth}
\usefont{T1}{ppl}{m}{n} % Palatino font
\small
% \color{MidnightBlue} % Using a deep blue color for the text
\textbf{Thought + Action 1}
\end{minipage}
}
}
\end{figure}

\begin{lstlisting}[language=Python]
THOUGHT 1: The detection tool did not annotate the balloons in the image. I will use the sliding_window_detection tool to search for the red and white balloons in the image.

ACTION 1:
```python
possible_patches, possible_boxes = sliding_window_detection(image_1, ["red balloon", "white balloon"])
for i, patch in enumerate(possible_patches):
    print(f"Patch {i+1}:")
    display(patch.annotated_image)
```
\end{lstlisting}

\begin{figure}[h]
\noindent\makebox[\textwidth][c]{
\framebox{%
\begin{minipage}{1.0\linewidth}
\usefont{T1}{ppl}{m}{n} % Palatino font
\small
% \color{MidnightBlue} % Using a deep blue color for the text
\textbf{Observation 1}

\end{minipage}
}
}
$\ \ \\$
Execution success. The output is as follows. 
\begin{minipage}{1.0\linewidth}
\usefont{T1}{ppl}{m}{n} % Palatino font
\small
Patch 1:
\end{minipage}

\includegraphics[width=0.45\linewidth]{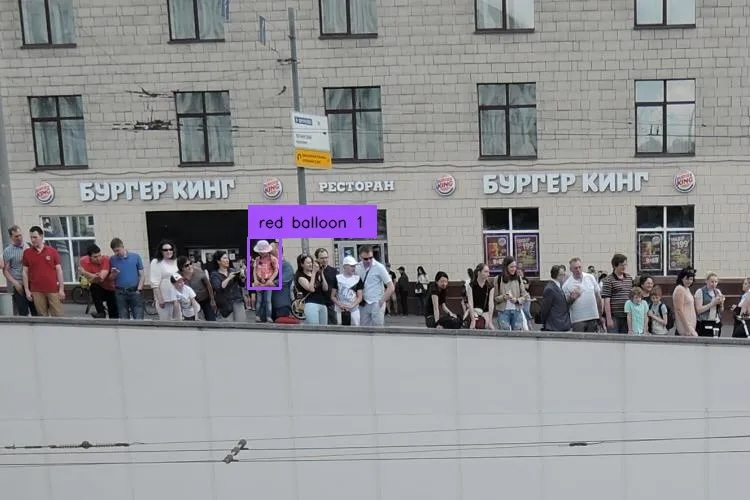}

\end{figure}
\begin{figure}

\begin{minipage}{1.0\linewidth}
\usefont{T1}{ppl}{m}{n} % Palatino font
\small
Patch 2:
\end{minipage}
\includegraphics[width=0.45\linewidth]{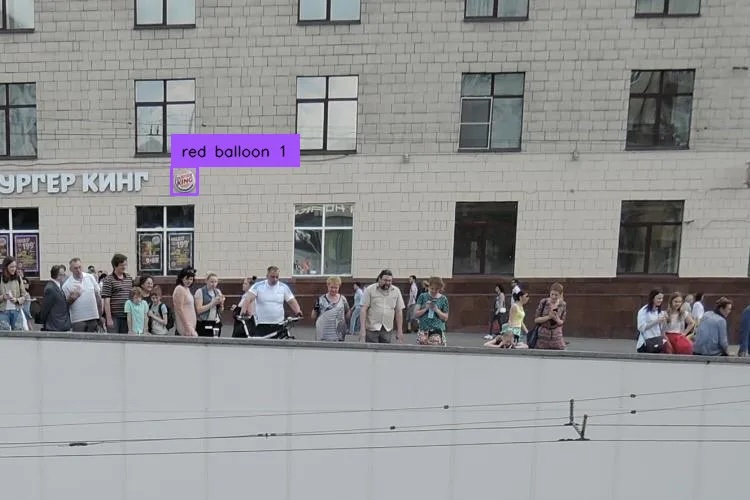}

\begin{minipage}{1.0\linewidth}
\usefont{T1}{ppl}{m}{n} % Palatino font
\small
Patch 3:
\end{minipage}
\includegraphics[width=0.45\linewidth]{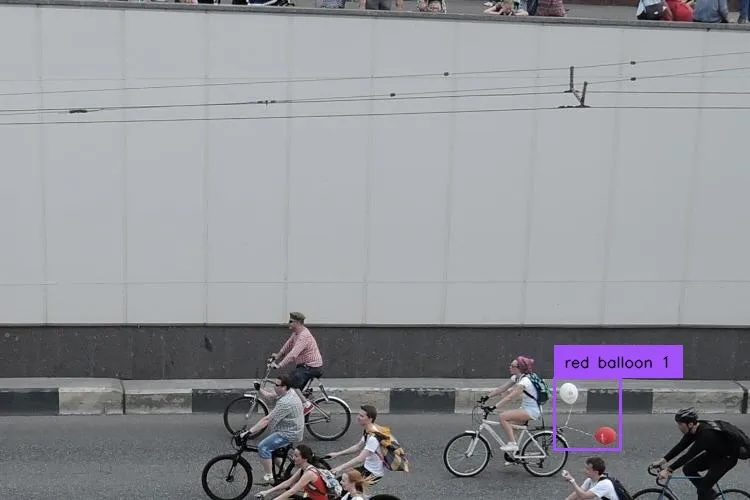}

\begin{minipage}{1.0\linewidth}
\usefont{T1}{ppl}{m}{n} % Palatino font
\small
Patch 4:
\end{minipage}
\includegraphics[width=0.45\linewidth]{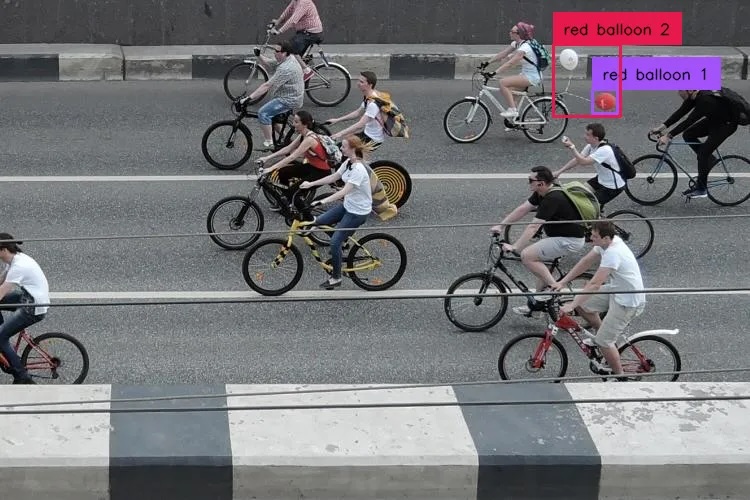}

\begin{minipage}{1.0\linewidth}
\usefont{T1}{ppl}{m}{n} % Palatino font
\small
Patch 5:
\end{minipage}
\includegraphics[width=0.45\linewidth]{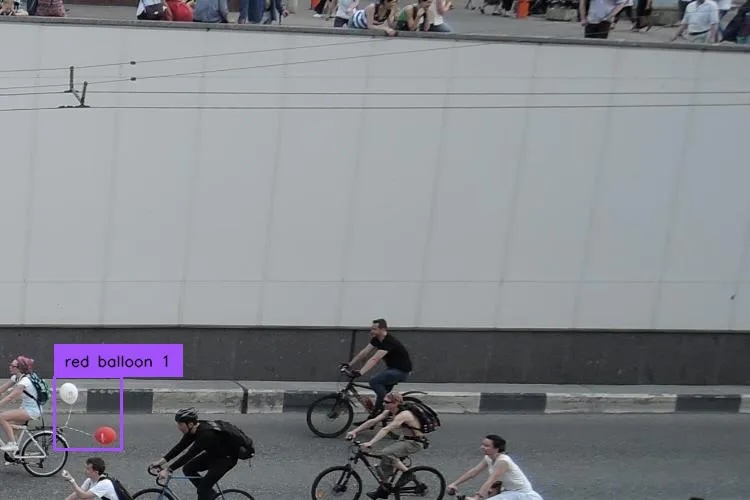}

\noindent\makebox[\textwidth][c]{

\begin{minipage}{1.0\linewidth}
\usefont{T1}{ppl}{m}{n} % Palatino font
\small
% \color{MidnightBlue} % Using a deep blue color for the text
Please generate the next THOUGHT and ACTION. If you can get the answer, please also reply with ANSWER: <your answer> and ends with TERMINATE.
\end{minipage}
}

\vspace{0.2in}
\noindent\makebox[\textwidth][c]{
\framebox{%
\begin{minipage}{1.0\linewidth}
\usefont{T1}{ppl}{m}{n} % Palatino font
\small
% \color{MidnightBlue} % Using a deep blue color for the text
\textbf{Thought + Action 2}
\end{minipage}
}
}
$\ \ \\$
THOUGHT 2: From Patch 4 and 5, I can see both the red and white balloons. The red balloon is clearly below the white balloon in these patches. 

ACTION 2: No action needed.

ANSWER: The red balloon is below the white balloon. So the answer is (A). TERMINATE
\end{figure}

\clearpage

% \weijia{
% - \textbf{Paper}:
% 1. add all prompts: add images 
% 2. compress images in latex

% - \textbf{Code}:
% 1. merge math into Yushi' code

\textbf{Prompts for math tasks.} The prompts are similar to the vision task for math tasks, except we remove the computer vision specialists, and add example codes for math plotting. Besides, the user query is different. For each task, the user query part of the prompt is as follows.
% as in Figures \ref{fig:bp-text-1} and \ref{fig:bp-text-2}.

% \lstinputlisting[language=Python]{prompt/math_convexity.py}
\begin{figure}[h]
\noindent\makebox[\textwidth][c]{
\framebox{%
\begin{minipage}{1.0\linewidth}
% \ttfamily
\usefont{T1}{ppl}{m}{n} % Palatino font
\footnotesize
\textbf{PROMPT}

You are given a real-valued, scalar function $f(x)$.

YOUR TASK is to determine whether $f(x)$ is an even function, an odd function, or neither.
Definition of an odd function: A function such that
$$f(-x) = -f(x)$$
where the sign is reversed but the absolute value remains the same if the sign of the independent variable is reversed.
A function is neither even nor odd if it does not satisfy either condition. 

Here is the expression of $f(x)$: $$f(x) = \frac{-2x^5}{2x^8 - 4x^6 + 12x^4 + 4x^2 + 11.16}$$

Respond with `even', `odd', `neither' first on whether the function $f(x)$ is even, odd, or neither, based on the definitions and your observation of the function. \hlwhite{You can generate matplotlib code to visualize the function.}

If you can get the answer, please reply with ANSWER: <your answer>, extract the final answer in FINAL ANSWER: <final answer> and ends with TERMINATE in the RESULT. 

\textit{Answer:
}
\end{minipage}
}
}
\caption{Prompt for the Math Parity task. We follow a similar prompt format to \cite{fu2024isobench}, except prompting the models to write the code to generate images. }
\label{fig:bp-text-2}
\end{figure}
\begin{figure}[h]
\noindent\makebox[\textwidth][c]{
\framebox{%
\begin{minipage}{1.0\linewidth}
% \ttfamily
\usefont{T1}{ppl}{m}{n} % Palatino font
\footnotesize
\textbf{PROMPT}

You are given a real-valued, scalar function $f(x)$.

YOUR TASK is to determine whether $f(x)$ is an convex function or concave function.
Definition of a convex function: A function such that 

$$\forall x, y, 0 \leq t \leq 1, 
f (tx + (1 - t)y) \leq t f (x) + (1 - t) f (y)$$

Definition of a concave function: A function such that 
$$\forall x, y, 0 \leq t \leq 1, 
f (tx + (1 - t)y) \geq t f (x) + (1 - t) f (y)$$

 Here is the expression of $f(x)$: $$f(x) = 7.57 - 0.08 * Abs(x) $$

Respond with `convex' or `concave' first on whether the function f (x) is convex or concave, based on the definitions and your observation of the function. \hlwhite{You can generate matplotlib code to visualize the function.} 

If you can get the answer, please reply with ANSWER: <your answer>, extract the final answer in FINAL ANSWER: <final answer> and ends with TERMINATE in the RESULT. 

\textit{Answer:
}
\end{minipage}
}
}
\caption{Prompt for the Math Convexity task. We follow the similar prompt format to \cite{fu2024isobench}, except prompting the models to write the code to generate images. }
\label{fig:bp-text-1}
\end{figure}
\begin{figure}[h]
\noindent\makebox[\textwidth][c]{
\framebox{%
\begin{minipage}{1.0\linewidth}
% \ttfamily
\usefont{T1}{ppl}{m}{n} % Palatino font
\footnotesize
% \color{MidnightBlue} % Using a deep blue color for the text
\textbf{PROMPT}

You are given an adjacency matrix of a graph and two query nodes. 

YOUR TASK is to find if there is a path between the two nodes.

\textit{Definition of connectivity:}
In an undirected graph G, two vertices u and v are called connected if G contains a path from u to v.
A path in a graph is a finite sequence of edges which joins a sequence of vertices.
In the query example, the nodes and the adjacency matrix are zero-indexed. 

\underline{Query Example:}

\textit{Adjacency Matrix: }

% [[0, 0, 0, 0, 0, 1, 0, 0, 0], 

%  [0, 0, 1, 0, 0, 0, 1, 0, 0], 
 
%  [0, 1, 0, 0, 1, 0, 0, 0, 0], 
 
%  [0, 0, 0, 0, 0, 0, 0, 0, 0], 
 
%  [0, 0, 1, 0, 0, 0, 0, 0, 0], 
 
%  [1, 0, 0, 0, 0, 0, 1, 0, 1], 
 
%  [0, 1, 0, 0, 0, 1, 0, 0, 0], 
 
%  [0, 0, 0, 0, 0, 0, 0, 0, 0], 
 
%  [0, 0, 0, 0, 0, 1, 0, 0, 0]]

\[
\begin{bmatrix}
0 & 0 & 0 & 0 & 0 & 1 & 0 & 0 & 0 \\
0 & 0 & 1 & 0 & 0 & 0 & 1 & 0 & 0 \\
0 & 1 & 0 & 0 & 1 & 0 & 0 & 0 & 0 \\
0 & 0 & 0 & 0 & 0 & 0 & 0 & 0 & 0 \\
0 & 0 & 1 & 0 & 0 & 0 & 0 & 0 & 0 \\
1 & 0 & 0 & 0 & 0 & 0 & 1 & 0 & 1 \\
0 & 1 & 0 & 0 & 0 & 1 & 0 & 0 & 0 \\
0 & 0 & 0 & 0 & 0 & 0 & 0 & 0 & 0 \\
0 & 0 & 0 & 0 & 0 & 1 & 0 & 0 & 0
\end{bmatrix}
\]

\textit{Query nodes indices (zero-indexed)}: 4 and 0

Respond with `yes' or `no' first on whether the query nodes are connected or not in the graph.

If there is a path, first provide the path as a sequence of vertices (nodes), and then explain your reasoning. \hlwhite{You can use networkx to draw the graph. }
If there is no path, explain why in details.
Answer (start with 'yes' or 'no'):
If you can get the answer, please reply with ANSWER: <your answer>, extract the final answer in FINAL ANSWER: <final answer> and ends with TERMINATE in the RESULT. 

\textit{Answer:}
\end{minipage}
}
}
\caption{Prompt for the Graph Connectivity task. We follow the similar prompt format to \cite{fu2024isobench}, except prompting the models to write the code to generate images.}
\label{fig:bp-text-2}
\end{figure}

\begin{figure}[h]
\noindent\makebox[\textwidth][c]{
\framebox{%
\begin{minipage}{1.0\linewidth}
\usefont{T1}{ppl}{m}{n} % Palatino font
\small
% \color{MidnightBlue} % Using a deep blue color for the text
\textbf{PROMPT}

You are given a visual representation of two graphs, graph G on the left and graph H on the right. 

YOUR TASK is to determine whether the two graphs are isomorphic to each other.

\textit{Definition of graph isomorphism:} In graph theory, an isomorphism of graphs G and H is a bijection \( f \) between the vertex sets of G and H, denoted as \( f: V(G) \rightarrow V(H) \). G and H are said to be isomorphic when \( f \) satisfies the following: any two vertices \( u \) and \( v \) of G are adjacent in G if and only if \( f(u) \) and \( f(v) \) are adjacent in H. This kind of bijection is commonly described as "edge-preserving bijection", in accordance with the general notion of isomorphism being a structure-preserving bijection.

In the query example, the adjacency matrices are zero-indexed.

\textit{Adjacency Matrix G: }
\[
\begin{bmatrix}
0 & 0 & 0 & 0 & 0 & 1 & 0 & 0 & 0 \\
0 & 0 & 1 & 0 & 0 & 0 & 1 & 0 & 0 \\
0 & 1 & 0 & 0 & 1 & 0 & 0 & 0 & 0 \\
0 & 0 & 0 & 0 & 0 & 0 & 0 & 0 & 0 \\
0 & 0 & 1 & 0 & 0 & 0 & 0 & 0 & 0 \\
1 & 0 & 0 & 0 & 0 & 0 & 1 & 0 & 1 \\
0 & 1 & 0 & 0 & 0 & 1 & 0 & 0 & 0 \\
0 & 0 & 0 & 0 & 0 & 0 & 0 & 0 & 0 \\
0 & 0 & 0 & 0 & 0 & 1 & 0 & 0 & 0
\end{bmatrix}
\]

\textit{Adjacency Matrix H: }
\[
\begin{bmatrix}
0 & 0 & 0 & 0 & 0 & 0 & 0 & 0 & 0 \\
0 & 0 & 1 & 0 & 0 & 0 & 1 & 0 & 0 \\
1 & 0 & 0 & 0 & 0 & 0 & 0 & 0 & 0 \\
0 & 0 & 0 & 0 & 0 & 0 & 0 & 0 & 0 \\
0 & 0 & 0 & 0 & 0 & 0 & 0 & 0 & 0 \\
1 & 0 & 0 & 0 & 0 & 0 & 1 & 0 & 0 \\
0 & 1 & 0 & 1 & 0 & 1 & 0 & 0 & 0 \\
0 & 0 & 0 & 0 & 0 & 0 & 0 & 1 & 0 \\
0 & 0 & 0 & 0 & 0 & 1 & 0 & 0 & 0
\end{bmatrix}
\]

Respond with 'yes' or 'no' first on whether the two graphs are isomorphic to each other. You can use networkx to draw the graph.
If they are isomorphic, first provide the bijection between the two graphs, and then explain your reasoning.
\hlwhite{You can use networkx to draw the graph. }
If they are not isomorphic, explain why in detail.
Answer (start with 'yes' or 'no'):
If you can get the answer, please reply with ANSWER: <your answer>, extract the final answer in FINAL ANSWER: <final answer> and ends with TERMINATE in the RESULT. 

\textit{Answer:}

\end{minipage}
}
}
\caption{Prompt for the Graph Isomorphism task. We follow a similar prompt format to \cite{fu2024isobench}, except prompting the models to write the code to generate images.}
\label{fig:bp-text-1}
\end{figure}
\begin{figure}[h]
\noindent\makebox[\textwidth][c]{
\framebox{%
\begin{minipage}{1.0\linewidth}
% \ttfamily
\usefont{T1}{ppl}{m}{n} % Palatino font
\footnotesize
% \color{MidnightBlue} % Using a deep blue color for the text
\textbf{PROMPT}

You are given an adjacency matrix of a graph and two query nodes (one source node and one sink node). The source node is the node where the flow starts and the sink node is the node where the flow ends.

YOUR TASK is to solve the maxflow problem given the weighted directed graph.

\textit{Definition of Maxflow problem:}
In the max flow problem, we have a directed graph with a source node \(s\) and a sink node \(t\), and each edge has a capacity (integer valued, colored in green) that represents the maximum amount of flow that can be sent through it.
The goal is to find the maximum amount of flow that can be sent from \(s\) to \(t\), while respecting the capacity constraints on the edges.

\underline{Query Example:}

\textit{Adjacency Matrix:}

% [[0, 1, 4], [0, 0, 6], [0, 0, 0]]

\[
\begin{bmatrix}
0 & 1 & 4 \\
0 & 0 & 6 \\
0 & 0 & 0
\end{bmatrix}
\]

\textit{Source node (zero-indexed)}: 0

\textit{Sink node (zero-indexed)}: 2

In the query example, the nodes and the adjacency matrix are zero-indexed. \hlwhite{You can use networkx to draw the graph. }
If you can get the answer, please reply with ANSWER: <your answer>, extract the final answer in FINAL ANSWER: <final answer> and ends with TERMINATE in the RESULT. 

% Use chain of thoughts to solve the question. Please don't write any code. Extract the final answer in FINAL ANSWER: <final answer>. Ends the conversation with TERMINATE.

\textit{Answer:}
\end{minipage}
}
}
\caption{Prompt for Graph Maxflow task. We follow the similar prompt format to \cite{fu2024isobench}, except prompting the models to solve the maxflow problem.}
\label{fig:bp-text-3}
\end{figure}
\begin{figure}[h]
\noindent\makebox[\textwidth][c]{
\framebox{%
\begin{minipage}{1.0\linewidth}
% \ttfamily
\usefont{T1}{ppl}{m}{n} % Palatino font
\footnotesize
\textbf{PROMPT}

Given the following FEN of the chess game:

1r1q1rk1/1b2b1Qp/4pp1B/pp1nP3/2pPN3/P1P5/1PB3PP/R4RK1 b - - 0 18	

Determine the game's outcome. Who won: White or Black? 
Answer (start with 'white' or 'black' or 'draw'):

\hlwhite{You can draw the chess board using Python given the FEN string.} 
If you can get the answer, please reply with ANSWER: <your answer>, extract the final answer in FINAL ANSWER: <final answer> and ends with TERMINATE in the RESULT. 

\textit{Answer:}

\end{minipage}
}
}
\caption{Prompt for Winner ID task. We follow a similar prompt format to \cite{fu2024isobench}, except prompting the models to analyze the game outcome. }
\label{fig:bp-text-1}
\end{figure}

\clearpage

\section{Dataset Statistics}
\label{sec:appendix:dataset_stat}

Table~\ref{tab:isobench} and ~\ref{tab:vision_task_stat} show the statistics of the datasets we used, including IsoBench~\cite{fu2024isobench}, BLINK~\cite{fu2024blink}, MMVP~\cite{tong2024eyes}, and $V^*$Bench~\cite{wu2023vstar}.

\begin{table*}[h]
\small
\centering
\vspace{-1em}
% {\fontsize{8pt}{11pt}
\selectfont
\begin{tabular}{l|cccc}
\toprule[1.2pt]
Dataset & size & partition & representation\\
\midrule
Math Parity & 383 & val & code \\
Math Convexity & 255 & val & code \\
Graph Maxflow & 128 & val & array \\
Graph Connectivity & 128 & val & array \\
Graph Isomorphism & 128 & val & array \\
Winner ID & 257 & val & FEN \\ 
\bottomrule[1.2pt]
\end{tabular}
% }
\vspace{-1ex}
\caption{IsoBench~\cite{fu2024isobench} data statistics.}
\vspace{-1ex}
\label{tab:isobench}
\end{table*}

\begin{table*}[h]
\small
\centering
\vspace{-1em}
% {\fontsize{8pt}{11pt}
\selectfont
\begin{tabular}{l|cccc}
\toprule[1.2pt]
Dataset & size & partition & input\\
\midrule
$V*$Bench & 257 & - & Single Image \\
MMVP & 300 & - & Single Image \\
BLINK Relative Depth & 124 & val & Single Image \\
BLINK Spatial Relation & 143 & val & Single Image \\
BLINK Jigsaw Puzzle & 150 & val  & Multiple Images \\
BLINK Visual Correspondence & 172 & val & Multiple Images \\
BLINK Semantic Correspondence & 139 & val & Multiple Image\\ 
\bottomrule[1.2pt]
\end{tabular}
% }
\vspace{-1ex}
\caption{Vision tasks data statistics.}
\vspace{-1ex}
\label{tab:vision_task_stat}
\end{table*}

\section{Costs}
\label{sec:appendix:cost}

The cost of running each task using GPT-4o is in Table~\ref{tab:cost}.
\begin{table*}[h]
\small
\centering
\vspace{-1em}
% {\fontsize{8pt}{11pt}
\selectfont
\begin{tabular}{l|ccc}
\toprule[1.2pt]
Dataset & tokens per sample &  GPT-4o cost per sample\\
\midrule
Math Parity & 2994 & \$0.015 \\
Math Convexity & 2211 & \$0.011 \\
Graph Connectivity & 2819 & \$0.014 \\
Graph Isomorphism & 3143 & \$0.016 \\
$V^*$Bench & 26647 & \$0.133 \\
MMVP & 11870 & \$0.059 \\
BLINK Relative Detph & 14078 & \$0.070 \\
BLINK Spatial Relation & 12848 & \$0.064 \\
BLINK Jigsaw Puzzle & 13206 & \$0.066 \\
BLINK Visual Correspondence & 16988 & \$0.085 \\
BLINK Semantic Correspondence & 11508 & \$0.058 \\
\bottomrule[1.2pt]
\end{tabular}
% }
\vspace{-1ex}
\caption{The cost of running \name on each task.}
\vspace{-1ex}
\label{tab:cost}
\end{table*}

% \name is implemented using the AutoGen~\cite{wu2023autogen} framework
% \noindent\textbf{Program execution in a Jupyter notebook environment.}
% LMs can sketch for multiple time steps to solve a task. For example, after adding an auxiliary line, the model might find that it needs to add another line. Taking this into consideration, different from prior work~\cite{gupta2023visual, surismenon2023vipergpt, hu2023visual, ma2024mms}, in which all codes are generated and executed once, \name uses a Jupyter notebook kernel-based program executor. In each time step, the LM has access to all prior executed code, variables, and objects. The model just needs to code for the next Notebook block, and we execute the new code block with the same kernel. This design reduces the amount of code the LM needs to generate since it can rely on past imports and variables.

% \paragraph{Debugging Based on Execution Feedback.} LMs sometimes generate erroneous code. Inspired by prior work on LM self-debugging~\cite{chen2023teaching}, when a code execution error occurs, the error message is included in the observation $o_{t+1}$. The model then refines its code based on this feedback in the next time step.

\section{Impact Statement}

Our work proposes \name, a framework aiming at advancing academic research and meeting industry needs. In a broader perspective, \name proposes a new way that humans can interact with LMs, and makes LMs more interpretable by eliciting their reasoning process with both language and sketches. On the other hand, if misused, the LMs may be used to generate harmful vision and text artifacts. Nevertheless, this is not directly related to our research, and more researchers can be involved to research on the safety issue in a multimodal context.

\clearpage
\section*{NeurIPS Paper Checklist}

\begin{enumerate}

\item {\bf Claims}
    \item[] Question: Do the main claims made in the abstract and introduction accurately reflect the paper's contributions and scope?
    \item[] Answer: \answerYes{}.
    \item[] Justification: Both the abstract and introduction reflect our paper's contributions and scope.

\item {\bf Limitations}
    \item[] Question: Does the paper discuss the limitations of the work performed by the authors?
    \item[] Answer: \answerYes{} % Replace by \answerYes{}, \answerNo{}, or \answerNA{}.
    \item[] Justification: We discuss the limitations of this work in \autoref{sec:conclusion}.

\item {\bf Theory Assumptions and Proofs}
    \item[] Question: For each theoretical result, does the paper provide the full set of assumptions and a complete (and correct) proof?
    \item[] Answer: \answerNA{} % Replace by \answerYes{}, \answerNo{}, or \answerNA{}.
    \item[] Justification: We do not make any theoretical claims.

    \item {\bf Experimental Result Reproducibility}
    \item[] Question: Does the paper fully disclose all the information needed to reproduce the main experimental results of the paper to the extent that it affects the main claims and/or conclusions of the paper (regardless of whether the code and data are provided or not)?
    \item[] Answer: \answerYes{} % Replace by \answerYes{}, \answerNo{}, or \answerNA{}.
    \item[] Justification: We provide implementation details in \S\ref{sec:method} and \S\ref{sec:appendix:prompt}.

\item {\bf Open access to data and code}
    \item[] Question: Does the paper provide open access to the data and code, with sufficient instructions to faithfully reproduce the main experimental results, as described in supplemental material?
    \item[] Answer: \answerNo{} % Replace by \answerYes{}, \answerNo{}, or \answerNA{}.
    \item[] Justification: The dataset is accessible online. We are committed to releasing the code.

\item {\bf Experimental Setting/Details}
    \item[] Question: Does the paper specify all the training and test details (e.g., data splits, hyperparameters, how they were chosen, type of optimizer, etc.) necessary to understand the results?
    \item[] Answer: \answerYes{} % Replace by \answerYes{}, \answerNo{}, or \answerNA{}.
    \item[] Justification: The test details are specified in \ref{sec:appendix:dataset_stat}, \ref{sec:experiments_math} and \ref{sec:experiments_vision}.

\item {\bf Experiment Statistical Significance}
    \item[] Question: Does the paper report error bars suitably and correctly defined or other appropriate information about the statistical significance of the experiments?
    \item[] Answer: \answerYes{} % Replace by \answerYes{}, \answerNo{}, or \answerNA{}.
    \item[] Justification: Since our work focus on inference of language models, the results are provided as evaluation scores on validation/test set. And we have explained how they are calculated in \ref{sec:experiments_math}.

\item {\bf Experiments Compute Resources}
    \item[] Question: For each experiment, does the paper provide sufficient information on the computer resources (type of compute workers, memory, time of execution) needed to reproduce the experiments?
    \item[] Answer: \answerYes{} % Replace by \answerYes{}, \answerNo{}, or \answerNA{}.
    \item[] Justification: We provide details of computing resources like compute device, number of workers, memory and time of execution in \S\ref{sec:appendix:cost}.
    
\item {\bf Code Of Ethics}
    \item[] Question: Does the research conducted in the paper conform, in every respect, with the NeurIPS Code of Ethics \url{https://neurips.cc/public/EthicsGuidelines}?
    \item[] Answer: \answerYes{} % Replace by \answerYes{}, \answerNo{}, or \answerNA{}.
    \item[] Justification: We have reviewed the NeurIPS Code of Ethics, and strictly followed it in this work.

\item {\bf Broader Impacts}
    \item[] Question: Does the paper discuss both potential positive societal impacts and negative societal impacts of the work performed?
    \item[] Answer: \answerYes{} % Replace by \answerYes{}, \answerNo{}, or \answerNA{}.
    \item[] Justification: Yes, we talk about boarder impacts in Appendix A.
    
\item {\bf Safeguards}
    \item[] Question: Does the paper describe safeguards that have been put in place for responsible release of data or models that have a high risk for misuse (e.g., pretrained language models, image generators, or scraped datasets)?
    \item[] Answer: \answerNA{} % Replace by \answerYes{}, \answerNo{}, or \answerNA{}.
    \item[] Justification: This paper focus on inference of language models, and thus there is no data or models to release.

\item {\bf Licenses for existing assets}
    \item[] Question: Are the creators or original owners of assets (e.g., code, data, models), used in the paper, properly credited and are the license and terms of use explicitly mentioned and properly respected?
    \item[] Answer: \answerYes{} % Replace by \answerYes{}, \answerNo{}, or \answerNA{}.
    \item[] Justification: Yes. We acknowledge the codebases we have referred to, cite and provide the urls of datasets and models in \autoref{sec:implementations}.

\item {\bf New Assets}
    \item[] Question: Are new assets introduced in the paper well documented and is the documentation provided alongside the assets?
    \item[] Answer: \answerNA{} % Replace by \answerYes{}, \answerNo{}, or \answerNA{}.
    \item[] Justification: This paper does not release new assets.

\item {\bf Crowdsourcing and Research with Human Subjects}
    \item[] Question: For crowdsourcing experiments and research with human subjects, does the paper include the full text of instructions given to participants and screenshots, if applicable, as well as details about compensation (if any)? 
    \item[] Answer: \answerNA{} % Replace by \answerYes{}, \answerNo{}, or \answerNA{}.
    \item[] Justification: This paper does not involve crowdsourcing nor research with human subjects.

\item {\bf Institutional Review Board (IRB) Approvals or Equivalent for Research with Human Subjects}
    \item[] Question: Does the paper describe potential risks incurred by study participants, whether such risks were disclosed to the subjects, and whether Institutional Review Board (IRB) approvals (or an equivalent approval/review based on the requirements of your country or institution) were obtained?
    \item[] Answer: \answerNA{} % Replace by \answerYes{}, \answerNo{}, or \answerNA{}.
    \item[] Justification: This paper does not involve crowdsourcing nor research with human subjects.

\end{enumerate}
% \clearpage
% \newpage
% \input{misc/checklist}
\end{document}